\documentclass[manuscript, screen, review=false]{acmart}
\AtBeginDocument{%
  }

\setcopyright{acmlicensed}
\copyrightyear{2018}
\acmYear{2018}
\acmDOI{XXXXXXX.XXXXXXX}

\acmJournal{JACM}
\acmVolume{37}
\acmNumber{4}
\acmArticle{111}
\acmMonth{8}

\usepackage{graphicx}

\usepackage{hyperref}       
\usepackage{xurl}            
\usepackage{booktabs}       
\usepackage{amsfonts}       
\usepackage{nicefrac}       
\usepackage{microtype}      
\usepackage{xcolor}         
\usepackage{enumitem}
\usepackage{amsmath}
\usepackage{amsthm}

\newtheorem{theorem}{Theorem}

\newtheorem{assumption}{Assumption}
\usepackage{algorithm}       

\usepackage{algpseudocode}  
\usepackage{amsmath}         
\usepackage{wrapfig}
\usepackage{amsfonts}        
\usepackage{listings}
\usepackage{xcolor}
\usepackage{multirow}
\usepackage{array}
\usepackage{tabularx}
\begin{document}

\title{Contexting as Recommendation: Evolutionary Collaborative Filtering for Context Engineering}


\author{Jiachen Zhu}
\authornote{Both authors contributed equally to this research.}
\email{gebro13@sjtu.edu.cn}
\affiliation{%
  \institution{Shanghai Jiao Tong University}
  \city{Shanghai}
  \country{China}
}

\author{Zhuoying Ou}
\authornotemark[1]
\email{zoeouzy23@sjtu.edu.cn}
\affiliation{%
  \institution{Shanghai Jiao Tong University}
  \city{Shanghai}
  \country{China}
}

\author{Congmin Zheng}
\email{desp.zcm@sjtu.edu.cn}
\affiliation{%
  \institution{Shanghai Jiao Tong University}
  \city{Shanghai}
  \country{China}
}

\author{Yuxiang Chen}
\email{yuxiang.chen.25@ucl.ac.uk}
\affiliation{%
  \institution{University College London}
  \city{London}
  \country{United Kingdom}
}

\author{Zeyu Zheng}
\email{zeyuzhen@andrew.cmu.edu}
\affiliation{%
  \institution{Carnegie Mellon University}
  \city{Pittsburgh}
  \state{Pennsylvania}
  \country{USA}
}

\author{Rong Shan}
\email{shanrong@sjtu.edu.cn}
\affiliation{%
  \institution{Shanghai Jiao Tong University}
  \city{Shanghai}
  \country{China}
}

\author{Lingyu Yang}
\email{jlnhbyu.yang@sjtu.edu.cn}
\affiliation{%
  \institution{Shanghai Jiao Tong University}
  \city{Shanghai}
  \country{China}
}

\author{Lionel Z. WANG}
\email{lionel-z.wang@connect.polyu.hk}
\affiliation{%
  \institution{The Hong Kong Polytechnic University}
  \city{Hong Kong}
  \country{China}
}

\author{Weiwen Liu}
\email{wwliu@sjtu.edu.cn}
\affiliation{%
  \institution{Shanghai Jiao Tong University}
  \city{Shanghai}
  \country{China}
}

\author{Yong Yu}
\email{yyu@sjtu.edu.cn}
\affiliation{%
  \institution{Shanghai Jiao Tong University}
  \city{Shanghai}
  \country{China}
}

\author{Weinan Zhang}
\email{wnzhang@sjtu.edu.cn}
\affiliation{%
  \institution{Shanghai Jiao Tong University}
  \city{Shanghai}
  \country{China}
}

\author{Jianghao Lin}
\correspondingauthor
\email{linjianghao@sjtu.edu.cn}
\affiliation{%
  \institution{Shanghai Jiao Tong University}
  \city{Shanghai}
  \country{China}
}
\renewcommand{\shortauthors}{Zhu et al.}

\begin{abstract}
Large Language Models (LLMs) are highly sensitive to their input contexts, motivating automated context engineering. Most existing methods optimize a single context strategy for average performance, implicitly assuming that the same context suits all inputs. However, theoretical intuition and our preliminary analysis reveal substantial heterogeneity in instance--context preferences, suggesting that global optimization leaves systematic gains unexplored.
We formulate context engineering as an inductive recommendation problem and introduce \textbf{Neural Collaborative Context Engineering (NCCE)}, which routes each input to a suitable strategy from a reusable context catalog. NCCE initializes diverse cluster-specific anchor contexts and then performs \textbf{Context-CF Co-Evolution}: a lightweight Neural Collaborative Filtering model learns instance--context compatibility to guide the evolution of specialized contexts, while newly evaluated contexts further refine the router. At inference time, the trained model assigns each unseen instance the context predicted to be most suitable. Theoretical analysis and experiments on three reasoning benchmarks show that NCCE consistently outperforms strong global context optimization baselines, demonstrating the value of instance-wise context routing.\end{abstract}

\ccsdesc[500]{Computing methodologies~Natural language processing}
\ccsdesc[500]{Information systems~Collaborative filtering}
\ccsdesc[300]{Information systems~Recommender systems}

\keywords{Context Engineering, Prompt Optimization, Collaborative Filtering, Large Language Models}

\received{20 February 2007}
\received[revised]{12 March 2009}
\received[accepted]{5 June 2009}

\maketitle
\section{Introduction}

\begin{wrapfigure}{R}{0.5\linewidth}
    \centering
    \includegraphics[width=\linewidth]{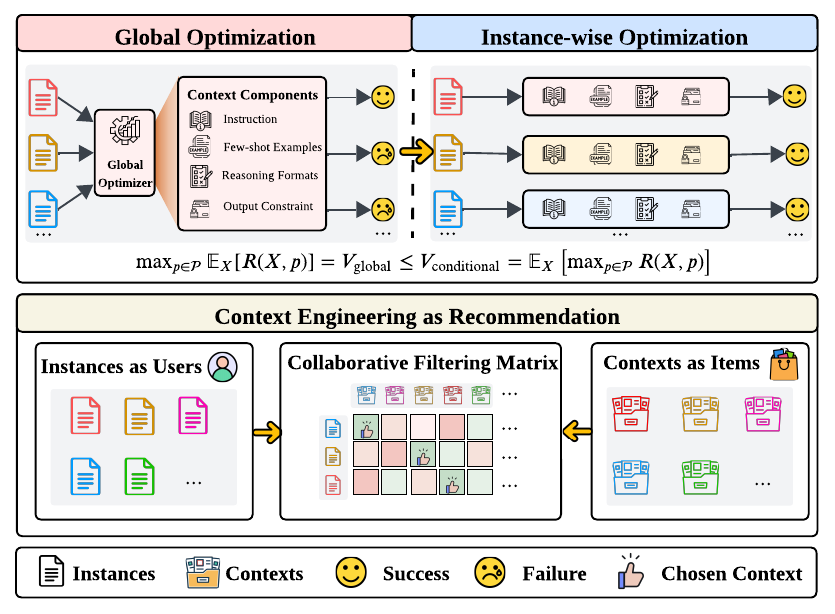}
    \caption{Context engineering as recommendation: learning to assign instance-specific composite contexts instead of optimizing a single global context strategy.}
    \label{fig:intro_fig}
\end{wrapfigure}

Large Language Models (LLMs) have demonstrated strong capabilities in complex reasoning, question answering, and context-dependent tasks~\cite{achiam2023gpt,touvron2023llama,brown2020language,wang2024openr}. However, their performance remains highly sensitive to the context provided at inference time. Small changes in instructions, few-shot demonstrations, reasoning formats, or output constraints can substantially alter model behavior. This sensitivity has motivated context engineering: the systematic design and optimization of input contexts for reliable LLM inference~\cite{zhao2021calibrate,khattab2023dspy,zhao2023survey}.

Most automated context optimization methods search for a single strategy that maximizes average performance over a dataset~\cite{APE,OPRO,textgrad,miprov2,evoprompt,gepa}. Although effective, this formulation implicitly assumes that one context is equally suitable for all inputs. In practice, context utility is instance-dependent: multi-hop questions may benefit from explicit decomposition, whereas verification tasks may require stricter evidence grounding.


This instance dependence also admits a simple decision-theoretic interpretation. Let $R(x,p)$ denote the reward of applying context strategy $p$ to instance $x$. The optimal input-conditioned policy has value $\mathbb{E}_x[\max_p R(x,p)]$, which is no smaller than that of the best global strategy, $\max_p\mathbb{E}_x[R(x,p)]$ (See Figure~\ref{fig:intro_fig}). The inequality becomes strict when different inputs prefer different contexts.

\begin{wrapfigure}{R}{0.5\linewidth}
    \centering
    \includegraphics[width=\linewidth]{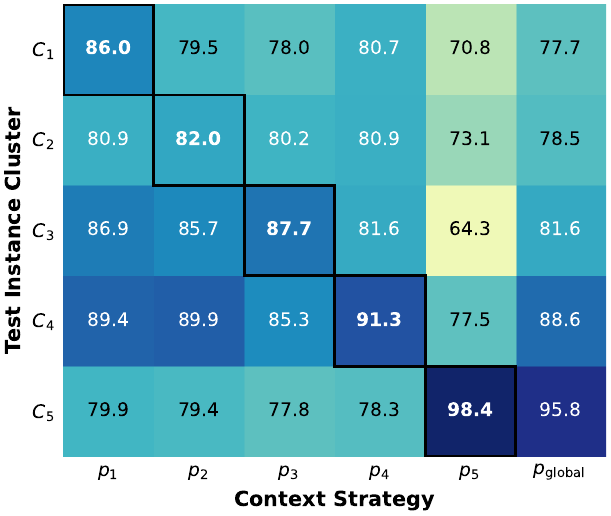}
    \caption{Cross-cluster performance of cluster-specific context strategies. Each row denotes a test cluster and each column a strategy optimized on the corresponding training cluster. Row-wise maxima lie on the diagonal, indicating cluster-dependent context preferences.}
    \label{fig:cross-cluster performance.}
\end{wrapfigure}





A preliminary study confirms that such preference heterogeneity arises in practice. We partition instances into five semantic clusters using K-Means and independently optimize one context strategy for each cluster with MIPROv2. Cluster-matched assignment achieves a macro-average score of 89.1, compared with 84.6 for the best single strategy across clusters (Figure~\ref{fig:cross-cluster performance.}). A qualitative example is also provided in Section~\ref{sec:case_study}.

The shift from a single global context to instance-level selection reshapes what context optimization should prioritize. A context with a strong average score may still provide only redundant coverage of instances already handled by the existing pool, whereas a context with a moderate average but complementary coverage can serve instances the current pool fails on. Determining which contexts are worth developing further therefore reduces to identifying which instances remain under-served, and instance-level execution feedback carries exactly this signal. Recent input-adaptive methods exploit such feedback to learn selectors over a fixed pool~\cite{wang2024one,petridis2024constitutionalexperts,sun2024query}. The same feedback can also drive the optimization itself, directing effort toward the instances where existing contexts still fall short.

Motivated by these observations, we formulate context engineering as an inductive recommendation problem~\cite{rendle2012bpr}. Input instances act as ``users,'' composite context strategies as ``items,'' and observed task outcomes as interaction signals. The objective is to learn a compatibility function that routes each unseen input to the most suitable strategy from a reusable context catalog. Under this view, global context optimization becomes a special case in which the routing function is constrained to select the same strategy for every input.

We introduce \textbf{Neural Collaborative Context Engineering (NCCE)} to jointly construct the context catalog and learn the routing function. NCCE first generates diverse cluster-specific anchor contexts to provide informative initial coverage. It then performs \textbf{Context-CF Co-Evolution}, in which a lightweight Neural Collaborative Filtering model~\cite{he2017neural} identifies unresolved instance regions and guides the evolution of specialized context variants. Newly evaluated contexts, in turn, provide additional interaction signals that refine the routing model. At inference time, the trained model scores the candidate strategies and assigns each unseen instance the context predicted to be most suitable.

The main contributions of this work are summarized as follows:
\begin{itemize}[leftmargin=12pt]
    \item We are the first to reframe context engineering from a global search problem into an instance-wise recommendation task, bridging decision-theoretic optimization with collaborative filtering to systematically address context-preference heterogeneity across inputs.
    
    \item We introduce \textit{Neural Collaborative Context Engineering}, featuring a novel \textit{Context-CF Co-Evolution} mechanism. By leveraging model latent gradients to expose catalog "blind spots," NCCE establishes a synergistic feedback loop where the context strategy catalog and the neural router iteratively refine each other.
    
    \item Extensive experiments validate that NCCE substantially outperforms both state-of-the-art global context optimizers and representative input-adaptive prompt-selection methods, demonstrating the value of jointly learning instance-level compatibility and evolving the context catalog.
\end{itemize}

\section{Preliminary}

We formalize composite context engineering and establish its connection to collaborative filtering.

\subsection{Context Engineering with Composite Strategies}
Let \(X=\{x_1,\dots,x_N\}\) denote a set of input instances and \(P=\{p_1,\dots,p_M\}\) a catalog of candidate context strategies. Each strategy is a composite configuration:
\[
p_j = \big(c^{\mathrm{inst}}_j,\; c^{\mathrm{demo}}_j,\; c^{\mathrm{reason}}_j,\; c^{\mathrm{out}}_j\big)
\]
representing the task instruction, few-shot demonstrations, reasoning format, and output constraints. Given a fixed LLM, the task accuracy of applying strategy \(p_j\) to instance \(x_i\) is:
\[
r_{ij}=R(x_i,p_j)
\]
where \(r_{ij}\in[0,1]\). Traditional optimization seeks a single globally optimal strategy:
\[
p^* = \arg\max_{p \in P} \sum_{i=1}^{N} R(x_i,p)
\]
This assumes one strategy fits all, ignoring the fine-grained interactions between specific instances and context formulations.

\subsection{Context Engineering as Recommendation}

To exploit instance heterogeneity, we reframe context engineering as a recommendation problem:

\textbf{Instances as users:} Each input instance \(x_i\in X\) is treated as a user whose semantic characteristics determine its preference over different context strategies.

\textbf{Context strategies as items.} Each context strategy \(p_j\in P\) is treated as an  item in the catalog.

\textbf{Accuracy as interaction.} The observed task accuracy \(r_{ij}\) obtained by applying \(p_j\) to \(x_i\) serves as the interaction signal between the instance and the context strategy.

Instead of a global optimum, we learn an instance-wise routing function to select the best context per input. Since evaluating all pairs is computationally prohibitive, we estimate compatibility from sparse observations.
\[
p_i^* = \arg\max_{p_j \in P} R(x_i,p_j)
\]
\textbf{Inductive Matrix Completion}
We model the sparse observed interactions $\Omega$ via inductive matrix completion. Instead of fixed IDs, we use semantic embeddings to predict compatibility:

$$ \hat{r}_{ij} = f_{\theta}\big(\phi(x_i), \psi(p_j)\big) $$

where $f_{\theta}$ scores the suitability between the embedded instance $\phi(x_i)$ and context $\psi(p_j)$. This inductive approach enables zero-shot routing for unseen instances and seamlessly integrates new contexts without retraining.

\section{Methodology}
\label{sec:methodology}

\begin{figure*}[t]
  \centering
  \includegraphics[width=1.0\textwidth]{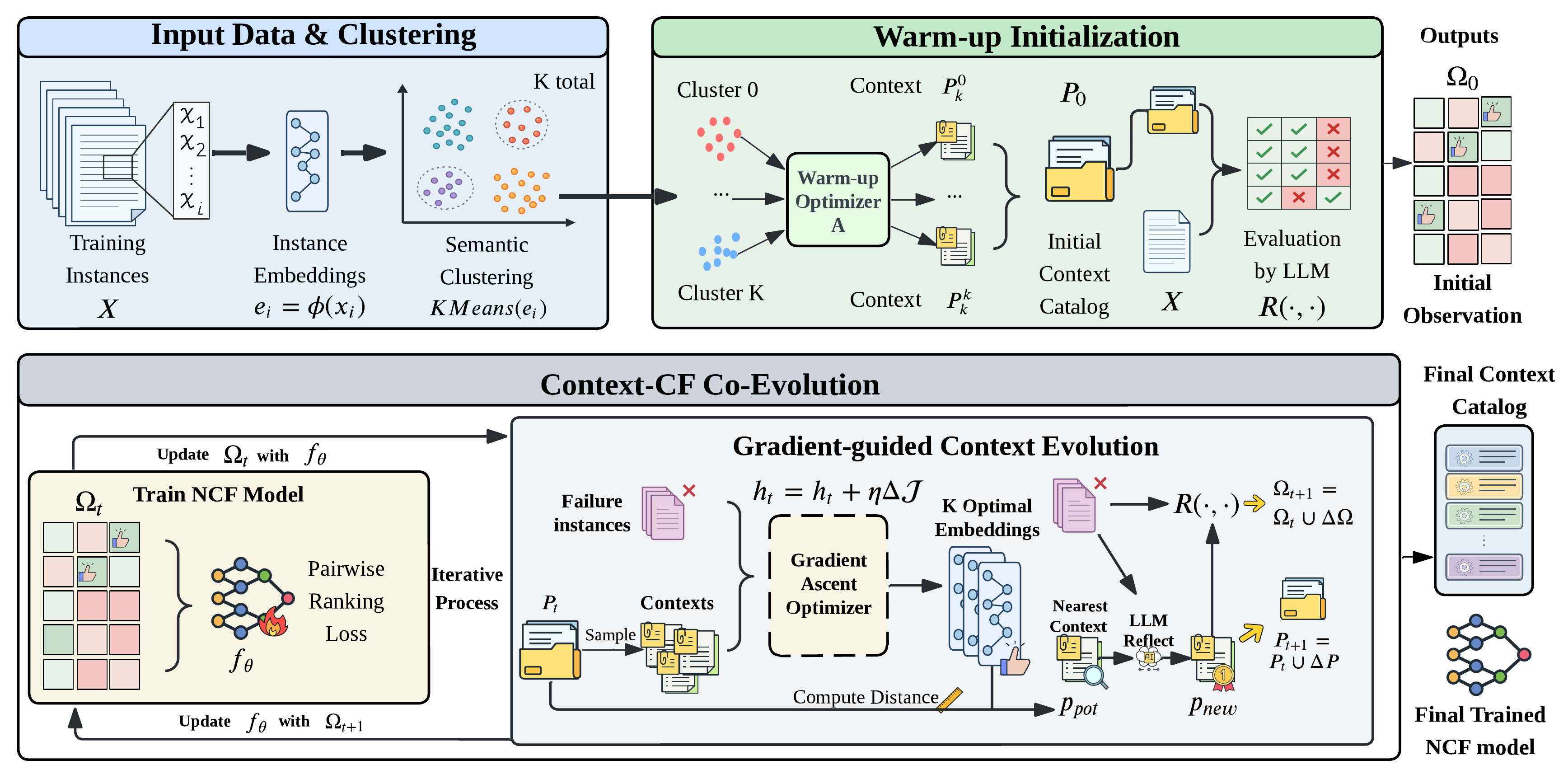}
  \caption{The overall architecture of NCCE, featuring a synergistic co-evolutionary loop between a neural collaborative filtering model and an evolving context catalog to enable personalized context construction and instance-wise routing.}
  \label{fig: Main Framework}
\end{figure*}

In this section, we present our NCCE framework. The framework operates in three stages: cluster-based initialization, Context-CF Co-Evolution, and instance-wise context routing. Algorithm~\ref{alg:ncce} summarizes the overall training procedure, which initializes a context catalog and iteratively updates both the catalog and the neural preference model.

\begin{algorithm*}[t]
\caption{Neural Collaborative Context Engineering (NCCE)}
\label{alg:ncce}
\begin{algorithmic}[1]
\Require Training instances \(X\), warm-up optimizer \(\mathcal{A}\), number of clusters \(K\), number of evolution iterations \(T\), LLM evaluator \(R(\cdot,\cdot)\)
\Ensure Final context catalog \(P_T\), trained preference model \(f_\theta\)

\State Embed training instances: \(e_i=\phi(x_i)\) for all \(x_i\in X\)
\State Partition instances into semantic clusters:
\[
\mathcal{C}_1,\ldots,\mathcal{C}_K=\mathrm{KMeans}(\{e_i\}_{i=1}^{N})
\]

\State Initialize context catalog \(P_0=\emptyset\)
\For{each cluster \(\mathcal{C}_k\)}
    \State Generate cluster-specific strategies \(P_k^0=\mathcal{A}(\mathcal{C}_k)\)
    \State Update catalog \(P_0 \leftarrow P_0 \cup P_k^0\)
\EndFor

\State Evaluate selected pairs from \(X\times P_0\) with the LLM evaluator \(R\)
\State Construct initial interaction set \(\Omega_0\)

\For{\(t=0,\ldots,T-1\)}
    \State Train preference model \(f_\theta\) on \(\Omega_t\) using pairwise ranking loss
    \State Use \(f_\theta\) to score candidate instance-context pairs
    \State Select seed contexts based on predicted suitability and diversity
    \State Generate new context strategies \(\Delta P_t\) via Algorithm~\ref{alg:gradient_context_evolution}: Gradient-guided Context Evolution
    \State Evaluate selected pairs from \(X\times \Delta P_t\) with \(R\), obtaining \(\Delta\Omega_t\)
    \State Update catalog: \(P_{t+1}\leftarrow P_t\cup \Delta P_t\)
    \State Update interactions: \(\Omega_{t+1}\leftarrow \Omega_t\cup \Delta\Omega_t\)
\EndFor

\State Train final preference model \(f_\theta\) on \(\Omega_T\)
\State \Return \(P_T, f_\theta\)

\end{algorithmic}
\end{algorithm*}

\subsection{Neural Collaborative Preference Model}

NCCE utilizes a lightweight Neural Collaborative Filtering (NCF) model to estimate instance-context suitability. Given an instance \(x_i\) and a context strategy \(p_j\), we extract their representations using frozen text encoders, where \(e_i=\phi(x_i)\) and \(h_j=\psi(p_j)\), and the composite context \(\psi(p_j)\) aggregates its individual components. We then project them into a shared latent space (\(u_i=W_x e_i, v_j=W_p h_j\)) and construct the interaction vector:
\[
z_{ij}=[u_i;\,v_j;\,u_i\odot v_j;\,|u_i-v_j|].
\]
The compatibility score is computed via an MLP:
\[
\hat{r}_{ij}=f_\theta(x_i,p_j)=\sigma(\mathrm{MLP}_\theta(z_{ij})).
\]

To optimize instance-wise selection, let \(\mathcal{D}_{\mathrm{pair}}\) denote the set of observed preference triples \((i,j,k)\), where \(p_j\) outperforms \(p_k\) on \(x_i\). NCCE minimizes the following pairwise ranking loss:
\[
\mathcal{L}_{\mathrm{rank}}
=
-\sum_{(i,j,k)\in\mathcal{D}_{\mathrm{pair}}}
\log \sigma(\hat{r}_{ij}-\hat{r}_{ik}).
\]
This objective mitigates the impact of inherent instance difficulty by focusing on relative ranking.

\subsection{Cluster-based Initialization}

An informative initial catalog is crucial for learning reliable preferences. To avoid weak starting points, NCCE employs cluster-based initialization. We partition the training instances into \(K\) clusters:
\[
\mathcal{C}_1,\ldots,\mathcal{C}_K
=
\mathrm{KMeans}(\{\phi(x_i)\}_{i=1}^{N}).
\]
For each cluster \(\mathcal{C}_k\), a warm-up optimizer \(\mathcal{A}\) generates specialized anchor contexts: \(P_k^{0}=\mathcal{A}(\mathcal{C}_k)\). The initial catalog becomes:
$
P_0=\bigcup_{k=1}^{K}P_k^{0}.
$ These diverse, group-level anchors provide much stronger preference learning signals than a single globally optimized strategy.

\subsection{Context-CF Co-Evolution}

Next, NCCE enters a co-evolutionary loop: the NCF model identifies promising directions for context improvement, while newly evolved contexts generate interaction data to refine the NCF model. 

At iteration \(t\), after training \(f_\theta\) on \(\Omega_t\), we target failure instances unsolved by any current strategy:
\[
\mathcal{F}_t = \{x_i \in X \mid R(x_i,p_j)=0,\ \forall p_j\in P_t\}.
\]
From \(\mathcal{F}_t\), we sample a batch \(\mathcal{B}_t\) of \(m\) failure instances. We then perform gradient ascent on \(k\) randomly sampled context embeddings \(h\) to maximize their NCF-predicted suitability for the failure batch:
\[
\mathcal{J}(h;\mathcal{B}_t) = \frac{1}{m} \sum_{x_i\in\mathcal{B}_t} s_\theta(h,x_i).
\]
The embedding is iteratively updated:
\[
h^{(\tau+1)} = \operatorname{Normalize} \left( h^{(\tau)} + \eta \nabla_{h^{(\tau)}} \mathcal{J}(h^{(\tau)};\mathcal{B}_t) \right)
\]
where \(\tau\) indexes the gradient-ascent steps, \(\eta\) is the learning rate, and \(\operatorname{Normalize}(\cdot)\) denotes \(\ell_2\)-normalization. 

After \(G\) steps, these optimized embeddings \(\tilde{h}_\ell\) represent continuous ideal contexts. To map them back to discrete text, we find the existing context \(p^{\mathrm{pot}}\) with the minimum average distance to these targets:
\[
p^{\mathrm{pot}} = \arg\min_{p_j\in P_t} \frac{1}{k} \sum_{\ell=1}^{k} \left\| \psi(p_j)-\tilde{h}_{\ell} \right\|_2.
\]
Through this \textit{gradient-guided selection}, NCCE avoids mutating blindly and instead selects the context nearest to the NCF's predicted optima. 

Finally, an LLM reflector revises \(p^{\mathrm{pot}}\) by diagnosing its failures on \(\mathcal{B}_t\), generating an improved strategy:
\[
p_{\mathrm{new}} = \mathcal{M}(p^{\mathrm{pot}}, \mathcal{B}_t).
\]
Evaluating \(p_{\mathrm{new}}\) yields new interactions \(\Delta\Omega_t\), updating both the catalog (\(P_{t+1}=P_t\cup\{p_{\mathrm{new}}\}\)) and the interaction set (\(\Omega_{t+1}=\Omega_t\cup\Delta\Omega_t\)) for the next NCF training round. Algorithm~\ref{alg:gradient_context_evolution} details this gradient-guided evolution procedure, from identifying failure instances and optimizing latent context embeddings to generating and evaluating a new context strategy.

\begin{algorithm*}[t]
\caption{Gradient-guided Context Evolution}
\label{alg:gradient_context_evolution}
\begin{algorithmic}[1]
\Require Current catalog $P_t$, observed interactions $\Omega_t$, trained NCF model $f_\theta$, evaluator $R$, failure batch size $m$, number of latent seeds $k$, gradient steps $G$, learning rate $\eta$, LLM reflector $\mathcal{M}$
\Ensure Updated catalog $P_{t+1}$, updated interactions $\Omega_{t+1}$

\State Train $f_\theta$ on $\Omega_t$ with pairwise ranking loss
\State Fix the parameters of $f_\theta$

\State Identify failure instances: $\mathcal{F}_t = \{x_i \in X \mid R(x_i,p_j)=0,\ \forall p_j\in P_t\}$
\State Sample $m$ failure instances: $\mathcal{B}_t \sim \mathcal{F}_t$

\State Randomly sample $k$ seed context strategies: $\{p_{a_1},\ldots,p_{a_k}\}\sim P_t$

\For{$\ell=1,\ldots,k$ \textbf{in parallel}}
    \State Initialize latent context embedding: $h_\ell^{(0)}=\psi(p_{a_\ell})$
    \For{$\tau=0,\ldots,G-1$}
        \State Compute failure-batch objective: $\mathcal{J}(h_\ell^{(\tau)};\mathcal{B}_t) = \frac{1}{m} \sum_{x_i\in\mathcal{B}_t} s_\theta(h_\ell^{(\tau)},x_i)$
        \State Update by gradient ascent: $h_\ell^{(\tau+1)} = \operatorname{Normalize} \left( h_\ell^{(\tau)} + \eta \nabla_{h_\ell^{(\tau)}} \mathcal{J}(h_\ell^{(\tau)};\mathcal{B}_t) \right)$
    \EndFor
    \State Set optimized target embedding: $\tilde{h}_\ell = h_\ell^{(G)}$
\EndFor

\For{each context strategy $p_j\in P_t$}
    \State Compute average distance to optimized targets: $d(p_j)= \frac{1}{k} \sum_{\ell=1}^{k} \left\| \psi(p_j)-\tilde{h}_\ell \right\|_2$
\EndFor

\State Select the most promising existing context: $p_{\mathrm{pot}} = \arg\min_{p_j\in P_t} d(p_j)$

\State Use LLM reflection to evolve $p_{\mathrm{pot}}$: $p_{\mathrm{new}} = \mathcal{M}(p_{\mathrm{pot}}, \mathcal{B}_t)$

\State Evaluate $p_{\mathrm{new}}$ on selected training instances and obtain $\Delta\Omega_t$

\State Update catalog: $P_{t+1}=P_t\cup\{p_{\mathrm{new}}\}$
\State Update interactions: $\Omega_{t+1}=\Omega_t\cup\Delta\Omega_t$

\State \Return $P_{t+1}, \Omega_{t+1}$

\end{algorithmic}
\end{algorithm*}


At inference time, the final context catalog and the trained NCF model are fixed. For each unseen instance, the model scores the candidate context strategies and selects the one with the highest predicted suitability.


\section{Theoretical Justification}
\label{sec:pac}

We provide a PAC-style analysis that justifies cluster-based initialization and clarifies its relationship to Context-CF Co-Evolution. The analysis decomposes the regret of instance-wise routing into two terms, each governed by a distinct component of NCCE.

Let $\mathcal{D}$ be the distribution over the input space $\mathcal{X}$ and let $P=\{p_1,\ldots,p_M\}$ be the context catalog. Denote by $r(x,p)\in[0,1]$ the true reward of applying context $p$ to instance $x$, and let $\hat{f}_\theta:\mathcal{X}\to P$ be the learned router. The instance-wise regret is
\[
\Delta(x) \;=\; \max_{p\in P}\, r(x,p) \;-\; r\bigl(x,\hat{f}_\theta(x)\bigr).
\]

Our analysis relies on a single structural assumption that links embedding geometry to context preferences. The remaining ingredients (anchor quality, cluster diameter, and anchor-violation Rademacher generalization) follow standard arguments and are detailed below.

\begin{assumption}[Cluster Lipschitz Preference]
\label{ass:cluster-lipschitz}
There exists a partition $\{\mathcal{C}_1,\ldots,\mathcal{C}_K\}$ of $\mathcal{X}$ and a constant $L>0$ such that for every cluster $\mathcal{C}_k$, every pair of instances $x,x'\in\mathcal{C}_k$, and every pair of contexts $p,p'\in P$,
\[
\bigl|\,(r(x,p)-r(x,p'))-(r(x',p)-r(x',p'))\,\bigr|
\;\le\;
L\,\|\phi(x)-\phi(x')\|.
\]
\end{assumption}

Assumption~\ref{ass:cluster-lipschitz} requires only that the \emph{relative} preference between two contexts varies smoothly within a cluster of semantically similar instances. It is strictly weaker than assuming Lipschitzness of $r(x,p)$ itself, and it aligns with the pairwise ranking objective: routing depends on relative rankings, not absolute reward magnitudes.

\begin{theorem}[Expected Regret Bound for Instance-wise Routing]
\label{thm:pac-bound}
Under Assumption~\ref{ass:cluster-lipschitz} and the bounded-loss and fixed-class conditions specified below, with probability at least $1-\delta$ over the draw of $n$ independent inputs $S=(X_1,\ldots,X_n)\sim\mathcal{D}^n$,
\[
\mathbb{E}_{x\sim\mathcal{D}}\!\bigl[\Delta(x)\bigr]
\;\le\;
\underbrace{\alpha + L\rho_K}_{\text{(I) catalog coverage}}
\;+\;
\underbrace{\widehat{\mathcal{R}}_n + 2\mathfrak{R}_n(\mathcal{H}) + \sqrt{\tfrac{\log(1/\delta)}{2n}}}_{\text{(II) router generalization}},
\]

where $\alpha=\max_k\alpha_k$ with $\alpha_k=\max_{p\in P}\mathbb{E}_{x|\mathcal{C}_k}[r(x,p)-r(x,p_k^0)]$ is the population optimality gap of the cluster anchor, $\rho_K$ is the cluster embedding diameter (satisfying $\rho_K=O(K^{-1/d})$ under regularity conditions on the instance distribution), $\widehat{\mathcal{R}}_n=\frac{1}{n}\sum_{i=1}^n\ell(x_i)$ is the empirical mean of the anchor-violation loss over the $n$ independent inputs (not the BPR training loss from the sparse interaction set $\Omega$), and $\mathfrak{R}_n(\mathcal{H})$ is the Rademacher complexity of the pre-fixed induced loss class $\mathcal{H}$.
\end{theorem}

\begin{proof}
We bound the expected regret of the trained router $\hat{f}_\theta$ on unseen test instances, recovering the form stated in Theorem~\ref{thm:pac-bound}.

We begin by analyzing a simpler reference router $f^{\mathrm{cluster}}$ that maps each instance $x\in\mathcal{C}_k$ to its cluster's anchor context $p_k^0$, and bound its expected regret within each cluster. Fix a cluster $\mathcal{C}_k$, any $x\in\mathcal{C}_k$, and any $p\in P$. By Assumption~\ref{ass:cluster-lipschitz} applied with $p'=p_k^0$, for any $x'\in\mathcal{C}_k$,
\[
r(x,p)-r(x,p_k^0) \;\le\; r(x',p)-r(x',p_k^0) + L\,\lVert\phi(x)-\phi(x')\rVert.
\]
Taking expectation over $x'\sim\mathcal{D}\,|\,\mathcal{C}_k$ and applying the cluster diameter bound $\lVert\phi(x)-\phi(x')\rVert\le\rho_K$ for all $x,x'\in\mathcal{C}_k$,
\[
r(x,p)-r(x,p_k^0)
\;\le\; \mathbb{E}_{x'|\mathcal{C}_k}\!\bigl[r(x',p)-r(x',p_k^0)\bigr] + L\rho_K.
\]
Maximizing over $p\in P$ and recalling $\alpha_k=\max_{p\in P}\mathbb{E}_{x'|\mathcal{C}_k}[r(x',p)-r(x',p_k^0)]$,
\[
\Delta^{\mathrm{cluster}}(x) \;\le\; \alpha_k + L\rho_K.
\]
Taking expectation over $x\sim\mathcal{D}\,|\,\mathcal{C}_k$ and marginalizing over clusters with $\alpha=\max_k\alpha_k$ yields a bound on the expected regret of the cluster-anchor router:
\begin{equation}
\label{eq:cluster-router-regret}
\mathbb{E}_{x\sim\mathcal{D}}\!\bigl[\Delta^{\mathrm{cluster}}(x)\bigr] \;\le\; \alpha+L\rho_K,
\end{equation}
where $\Delta^{\mathrm{cluster}}(x)=\max_p r(x,p) - r(x,f^{\mathrm{cluster}}(x))$.

We now bound the gap between the trained router $\hat{f}_\theta$ and the cluster-anchor reference. Let $a(x)=p_k^0$ for $x\in\mathcal{C}_k$. Define the \emph{anchor-violation loss}
\[
\ell(x) \;=\; \max_{q\in P}\,\bigl[r(x,a(x))-r(x,q)\bigr]_+\,
\mathbf{1}\!\bigl\{s_\theta(x,q)\ge s_\theta(x,a(x))\bigr\},
\]
measuring the maximal reward lost when the scorer $s_\theta$ ranks a suboptimal context $q$ at least as high as the anchor. Since any additional regret of $\hat{f}_\theta$ beyond $f^{\mathrm{cluster}}$ arises from exactly such a misordering,
\[
\Delta(x) \;\le\; \Delta^{\mathrm{cluster}}(x)+\ell(x).
\]
Taking expectations and applying a standard uniform Rademacher generalization bound~\citep{bartlett2002rademacher} to $\ell$ over the hypothesis class $\mathcal{H}$ (jointly covering the catalog, anchors, and router), with probability at least $1-\delta$,
\[
\mathbb{E}[\ell(X)] \;\le\; \widehat{\mathcal{R}}_n + 2\mathfrak{R}_n(\mathcal{H}) + \sqrt{\tfrac{\log(1/\delta)}{2n}}.
\]
Combining with~\eqref{eq:cluster-router-regret} gives
\begin{equation}
\label{eq:expected-regret}
\mathbb{E}_{x\sim\mathcal{D}}\!\bigl[\Delta(x)\bigr] \;\le\; \alpha+L\rho_K+\widehat{\mathcal{R}}_n+2\mathfrak{R}_n(\mathcal{H})+\sqrt{\tfrac{\log(1/\delta)}{2n}}.
\end{equation}

Equation~\eqref{eq:expected-regret} is the form stated in Theorem~\ref{thm:pac-bound}.
\end{proof}

Theorem~\ref{thm:pac-bound} decomposes the routing regret into two terms that are addressed by complementary mechanisms in NCCE. Term~(I) is primarily governed by the initial catalog: $L\rho_K$ shrinks as the number of clusters $K$ increases, and $\alpha$ reflects how closely each cluster anchor approximates the best available context for that cluster in expectation. This term cannot be reduced by enlarging the interaction set $\Omega$ alone without also improving anchor quality, and it is precisely what cluster-based initialization is designed to control. The contrast with global context optimization follows from two independent observations. First, a globally optimized strategy ($K=1$) sets Term~(I) to $L\cdot\mathrm{diam}(\phi(\mathcal{X}))+\alpha$, which is large whenever instance preferences vary across the full embedding space, showing that the approximation term in our bound grows with the semantic diameter. Second, and separately, whenever the instance distribution is heterogeneous—meaning every fixed context fails to achieve near-optimal reward on a non-trivial fraction of instances—any global strategy incurs strictly positive expected regret as a direct consequence of heterogeneity; this lower bound on global-strategy regret is independent of the upper bound in Theorem~\ref{thm:pac-bound}.

Term~(II) is primarily governed by the size and informativeness of the interaction set and follows the standard form of Rademacher generalization applied to the anchor-violation loss. This is the term that Context-CF Co-Evolution is designed to reduce, by iteratively expanding $\Omega$ with new instance-context evaluations. Because the two terms address different bottlenecks, cluster-based initialization and Context-CF Co-Evolution play complementary rather than redundant roles in the NCCE pipeline. Note that $\mathcal{H}$ is fixed before the bound is applied and must cover all loss functions that the algorithm might select; catalog evolution can change $\alpha$ (since $\alpha$ depends on $P$) and shift which element of $\mathcal{H}$ is active, but $\mathcal{H}$ itself remains fixed within any single application of the bound.

The bound also suggests a possible finite-sample trade-off with respect to $K$: increasing $K$ reduces $L\rho_K$ but may increase $\alpha$ when available training instances per cluster are limited, degrading anchor quality. The best $K$ is therefore dataset-dependent and should be selected to minimize the combined bound. 

\paragraph{Remarks.}
The expected regret bound already exposes the qualitative dependence on $K$, $\alpha$, and $L\rho_K$ that motivates the design of NCCE. A tail-probability form can be recovered by applying Markov's inequality, introducing a $1/\varepsilon$ factor on both terms. The catalog size $M$ may also affect $\mathfrak{R}_n(\mathcal{H})$, since the loss class must cover all catalog choices; this explains why expanding the catalog without growing the interaction set $\Omega$ does not by itself reduce regret. Finally, $\widehat{\mathcal{R}}_n$ is the theoretical anchor-violation loss averaged over all $n$ inputs; it is not directly equal to the BPR ranking loss computed from the sparse interaction set $\Omega$. Expressing the bound purely in terms of observed pairs would require an additional coverage residual, which we leave to future work.
\section{Experiments}
\label{sec:experiment}

We evaluate whether NCCE improves task accuracy by constructing instance-specific contexts. Our code is publicly available.~\footnote{\url{https://anonymous.4open.science/r/Context_Engineering_Collaborative_Filtering-1EF5}}

\subsection{Experimental Setup}




\textbf{Datasets and Metrics.}
We evaluate NCCE on three reasoning-oriented benchmarks: HoVer, SCONE, and HotpotQA. These datasets cover different forms of instance heterogeneity and therefore provide a suitable testbed for instance-wise context construction.

\paragraph{HoVer.}~\cite{jiang2020hover}
HoVer is a multi-hop fact verification benchmark. Each instance requires verifying a claim based on evidence that may span multiple documents. This task benefits from context strategies that encourage evidence grounding, decomposition, and careful verification.

\paragraph{SCONE.}~\cite{she2023scone}
ScoNe is a negation reasoning benchmark. Each instance consists of a sentence pair in which the hypothesis involves negation over category or set-membership relations, and the task is to determine whether the negated statement follows from the premise. This task benefits from context strategies that help the model reason carefully about negation scope and category entailment.

\paragraph{HotpotQA.}~\cite{yang2018hotpotqa}
HotpotQA is a multi-hop question answering benchmark. Each instance requires reasoning over multiple pieces of evidence to produce the final answer. This task is useful for testing whether NCCE can route different questions to context strategies with appropriate reasoning formats and demonstrations.

Across all datasets, we use task accuracy as the primary evaluation metric. For each instance, the prediction is considered correct if the final answer matches the ground-truth label or answer under the dataset-specific evaluation protocol.

\textbf{Data Partitioning and Statistics.}
To evaluate the effectiveness of the instance-wise context routing, we adopt a systematic partitioning strategy for all three datasets. 

\paragraph{Data Splitting Protocol} For each dataset, we first aggregate all available samples and perform a random shuffle. A fixed number of instances are then held out as a Test Set to ensure an unbiased final evaluation. The remaining samples form a Training/Dev Pool. For ScoNe, which organizes instances into contrastive groups of six variants sharing the same base sentence, the train+dev/test split is performed at the group level so that all variants of the same source sentence fall entirely within either the train+dev pool or the test set; no variant-level leakage across the split boundary is possible. Within the train+dev pool, individual instances are then treated independently for the subsequent cluster-based partitioning step described below.

\paragraph{Cluster-based Partitioning} To facilitate the Context-CF co-evolution, we apply K-means clustering to this pool based on the input text embeddings. Within each cluster, samples are further divided into Training and Dev sets using a 1:1 ratio. This cluster-aware split ensures that both the router training and the context evolution stages have balanced access to the diverse semantic patterns identified during the clustering phase.

\paragraph{Dataset Statistics} Table~\ref{tab:data_budget} summarizes the dataset partitions.

\begin{table}[t]
\centering
\caption{Dataset partitions and estimated total API calls.
API counts include bootstrapping, evolutionary rounds, and final evaluation.}
\label{tab:data_budget}
\begin{tabular}{lrrrrr}
\toprule
Dataset & Total & Train & Dev & Test & API Calls \\
\midrule
HoVer    & 2,520 & 500 & 500 & 1,520 & $\sim$40,000  \\
SCONE    & 2,200 & 500 & 500 & 1,200 & $\sim$105,000 \\
HotpotQA & 3,000 & 500 & 500 & 2,000 & $\sim$16,000  \\
\bottomrule
\end{tabular}
\end{table}

\textbf{Implementation Details.}
We employ specialized models for different roles within the NCCE framework. Specifically, we use GPT-4o-mini as the task model to perform reasoning and generate final answers. The context catalog is initialized via semantic clustering and DSPY~\cite{miprov2}-based warm-up optimization. For the Context-CF Co-Evolution stage, GPT-4o \cite{openai2024gpt4ocard} is utilized as the prompt model to refine and generate context variants. Additionally, all-MiniLM-L6-v2 \cite{wang2020minilm} is adopted as the embedding model to compute semantic features for clustering and NCF router input.

For semantic encoding, we use a frozen text encoder to represent both input instances and context strategies. The NCF model is trained with the pairwise ranking loss described in Section~\ref{sec:methodology}. Unless otherwise specified, NCCE uses \(K\) semantic clusters for initialization and performs \(T\) rounds of context-CF co-evolution.

The warm-up optimizer in cluster-based initialization is MIPROv2. We emphasize that MIPROv2 plays two roles in our experiments: it serves as a strong global optimization baseline, and it is also used as a replaceable local optimizer for generating cluster-specific anchor contexts in NCCE.

All large language model components are accessed via official OpenAI APIs. During the training and testing phases, we recorded the frequency of API interactions. Table~\ref{tab:data_budget} provides an estimated total number of API calls for each dataset, encompassing bootstrapping, evolutionary rounds, and final evaluation.

The training and optimization of the MLP-based NCF router are conducted on a single NVIDIA RTX 4090 GPU. Given the lightweight design of our router, this setup provides sufficient computational power for rapid co-evolution and efficient inference. 

\textbf{Router Training Configurations.}
Here we provide the detailed hyperparameter search space and final configurations for router training. The router is implemented as a Multi-Layer Perceptron with two hidden layers of 1024 and 512 units, respectively, using ReLU as the activation function. For each evolution round, we perform a grid search over the following candidate values: Learning Rate $\eta \in \{1\text{e-}2, 5\text{e-}3, 2\text{e-}3, 1\text{e-}3, 5\text{e-}4, 1\text{e-}4\}$, Batch Size $m \in \{64, 128, 256, 512\}$, Dropout $\in \{0.05, 0.1, 0.15, 0.2\}$, Temperature $\tau \in \{0.8, 1.0, 1.2\}$, and Lambda $\lambda \in \{1\text{e-}1, 5\text{e-}2, 1\text{e-}2, 5\text{e-}3, 1\text{e-}3, 5\text{e-}4\}$. The optimal hyperparameters selected for each evolution round across the three datasets are summarized in Table~\ref{tab:router_configs}.

\begin{table}[tbp]
\centering
\setlength{\tabcolsep}{4pt}
\renewcommand{\arraystretch}{0.95}
\caption{Router training configurations across evolution rounds. All configurations use a maximum of 30 epochs with an early-stopping patience of 5.}
\label{tab:router_configs}
\begin{tabular}{lcccccc}
\toprule
\textbf{Dataset} & \textbf{Round} & \textbf{LR} & \textbf{Batch Size} & \textbf{Dropout} & \textbf{Temp.} & $\boldsymbol{\lambda}$ \\
\midrule
\multirow{6}{*}{HoVer}
 & 0 & 5e-3 & 64  & 0.10 & 1.0 & 1e-2 \\
 & 1 & 2e-3 & 512 & 0.10 & 1.0 & 5e-2 \\
 & 2 & 5e-3 & 256 & 0.10 & 1.0 & 1e-1 \\
 & 3 & 2e-3 & 256 & 0.10 & 1.2 & 1e-1 \\
 & 4 & 1e-3 & 512 & 0.10 & 1.0 & 1e-1 \\
 & 5 & 2e-3 & 64  & 0.05 & 1.2 & 1e-3 \\
\midrule
\multirow{6}{*}{SCONE}
 & 0 & 5e-3 & 128 & 0.10 & 0.8 & 1e-3 \\
 & 1 & 5e-3 & 64  & 0.05 & 1.2 & 5e-4 \\
 & 2 & 5e-3 & 64  & 0.15 & 1.0 & 1e-3 \\
 & 3 & 5e-3 & 64  & 0.05 & 1.2 & 5e-4 \\
 & 4 & 5e-3 & 128 & 0.10 & 1.0 & 1e-3 \\
 & 5 & 1e-2 & 256 & 0.10 & 1.0 & 1e-3 \\
\midrule
\multirow{6}{*}{HotpotQA}
 & 0 & 1e-3 & 128 & 0.10 & 1.2 & 1e-2 \\
 & 1 & 1e-3 & 128 & 0.10 & 0.8 & 1e-2 \\
 & 2 & 1e-3 & 128 & 0.10 & 1.2 & 1e-2 \\
 & 3 & 5e-4 & 64  & 0.10 & 1.0 & 1e-2 \\
 & 4 & 1e-4 & 64  & 0.10 & 0.8 & 1e-3 \\
 & 5 & 5e-4 & 256 & 0.20 & 1.2 & 5e-3 \\
\bottomrule
\end{tabular}
\end{table}

\textbf{Baselines.}
We compare NCCE with APE~\cite{APE}, OPRO~\cite{OPRO}, EvoPrompt~\cite{evoprompt}, TextGrad~\cite{textgrad}, GEPA~\cite{gepa}, and MIPROv2~\cite{miprov2}, which all optimize a single global context strategy. We further include instance-adaptive baselines MoP~\cite{wang2024one}, Prompt-OIRL~\cite{sun2024query}, and ConstitutionalExperts~\cite{petridis2024constitutionalexperts}, which route inputs across a fixed context catalog.
For baselines with configurable iteration or candidate counts, we set the number of GPT-4o and GPT-4o-mini calls to match NCCE's total API budget; baselines that do not expose such parameters are run with their default settings.

\paragraph{APE.}
Automatic Prompt Engineer generates and selects task instructions from a pool of LLM-proposed candidates, optimizing the instruction according to a task-specific score.

\paragraph{OPRO.}
Optimization by PROmpting uses an LLM as a black-box optimizer. It iteratively conditions on previous candidate prompts and their scores to generate improved prompts.

\paragraph{EvoPrompt.}
EvoPrompt connects LLMs with evolutionary algorithms. It maintains a population of prompts and applies evolutionary operators to generate new candidates, which are selected according to development-set performance.

\paragraph{TextGrad.}
TextGrad optimizes textual variables by backpropagating natural-language feedback produced by LLMs. We use it to optimize the textual components of the context strategy.

\paragraph{MIPROv2.}
MIPROv2 optimizes instructions and few-shot demonstrations for downstream task performance. In our experiments, MIPROv2 serves both as a strong global optimization baseline and as the replaceable warm-up optimizer used in NCCE's cluster-based initialization.

\paragraph{GEPA.}
GEPA is a reflective prompt evolution method that uses LLM-generated reflections over execution traces to propose prompt updates and selects candidates through Pareto-aware evolutionary search.

\textit{GEPA-Merge.} GEPA-Merge augments GEPA with a system-aware crossover operator. It combines complementary module-level prompt updates from candidate systems that share a common ancestor, proposing new candidates alongside reflective mutation.

\textit{OpenEvolve.} OpenEvolve is an open-source evolutionary optimization framework applicable to programs and prompts. It uses LLM-generated modifications and task-specific evaluation feedback to evolve diverse candidate populations through MAP-Elites and island-based search.

\textit{POLCA.} POLCA formulates prompt and system optimization as stochastic generative search guided by numerical rewards and textual feedback. It combines a priority queue, an $\varepsilon$-net diversity mechanism, and LLM-based summarization of historical trials to guide candidate selection and generation.

\paragraph{MoP.}
Mixture-of-Prompts constructs a set of expert prompts, each combining task instructions and few-shot demonstrations, by running a prompt optimizer independently within each semantic cluster of training instances. At inference time, each query is routed to the expert whose cluster centroid is nearest in the embedding space.

\paragraph{ConstitutionalExperts.}
ConstitutionalExperts organizes training inputs into semantic clusters and iteratively refines a principle-based prompt expert for each cluster using failure feedback. Routing at inference time assigns each query to the nearest cluster centroid.

\paragraph{Prompt-OIRL.}
Prompt-OIRL trains a query-dependent reward model from offline prompt execution outcomes via inverse reinforcement learning. At inference, the reward model scores all candidate prompts for a given query, and the top-ranked prompt is selected.

\begin{table*}[t]
\centering
\label{-50pt}
\caption{Main accuracy results on HoVer, SCONE, and HotpotQA. Baselines include global context optimization methods and instance-adaptive methods that route over a fixed prompt catalog. NCCE performs instance-wise context routing over a co-evolved catalog. The best results are shown in \textbf{bold} and the second best results are \underline{underlined}. $^*$ means $p-value<0.05$ in significance test.}
\begin{tabular}{lcccccccc}
\toprule
\textbf{Method} & \multicolumn{2}{c}{\textbf{HoVer}} & \multicolumn{2}{c}{\textbf{SCONE}} & \multicolumn{2}{c}{\textbf{HotpotQA}} & \multicolumn{2}{c}{\textbf{Average}} \\
 & dev & test & dev & test & dev & test & dev & test \\
\midrule
zero-shot & 67.5 & 68.2 & 70.6 & 70.5 & 37.0 & 33.4 & 58.4 & 57.4 \\
fixed few-shot & 71.4 & 70.9 & 78.8 & 79.6 & 57.2 & 51.6 & 69.1 & 67.4 \\
APE~\cite{APE} & 71.4 & 69.8 & 78.0 & 78.1 & 54.8 & 51.7 & 68.1 & 66.5 \\
OPRO~\cite{OPRO} & 73.4 & 70.9 & 73.0 & 75.6 & 37.6 & 34.5 & 61.3 & 60.3 \\
EvoPrompt~\cite{evoprompt} & 72.8 & 70.5 & 76.4 & 78.7 & 51.8 & 50.3 & 67.0 & 66.5 \\
TextGrad~\cite{textgrad} & 72.1 & 71.2 & 72.4 & 74.5 & 48.2 & 44.0 & 64.2 & 63.2 \\
MIPROv2~\cite{miprov2} & 73.9 & 71.6 & \underline{85.6} & 83.5 & 56.0 & 51.2 & 71.8 & 68.8 \\
GEPA~\cite{gepa} & 71.1 & 69.9 & 84.2 & 84.2 & 52.0 & 49.3 & 69.1 & 67.8 \\
GEPA-Merge~\cite{gepa} & 72.1 & 70.4 & 84.4 & \underline{86.2} & 53.2 & 49.6 & 69.9 & 68.7 \\
OpenEvolve~\cite{openevolve} & \textbf{76.1} & \underline{73.8} & 79.4 & 79.2 & 56.2 & 53.1 & 70.6 & 68.7 \\
POLCA~\cite{ren2026polca} & 72.5 & 70.2 & 83.2 & 85.8 & \underline{63.8} & \underline{58.6} & \underline{73.2} & \underline{71.5} \\
MoP~\cite{wang2024one} & 67.5 & 71.0 & 82.2 & 78.9 & 63.4 & 54.7 & 71.0 & 68.2 \\
Prompt-OIRL~\cite{sun2024query} & 72.0 & 72.5 & 80.0 & 83.4 & 62.0 & 55.5 & 71.3 & 70.5 \\
ConstitutionalExperts~\cite{petridis2024constitutionalexperts} & 72.8 & 68.8 & 77.4 & 72.7 & 60.2 & 57.0 & 70.1 & 66.2 \\
NCCE (Ours) & \underline{75.8} & \textbf{74.7$^*$} & \textbf{93.4$^*$} & \textbf{89.7$^*$} & \textbf{69.8$^*$} & \textbf{60.1$^*$} & \textbf{79.7$^*$} & \textbf{74.8$^*$} \\
\bottomrule
\end{tabular}
\label{tab:main_results}
\end{table*}

\subsection{Main Results}

Table~\ref{tab:main_results} reports the main accuracy results. Across all three datasets, NCCE consistently and significantly improves over all baselines, validating our core hypothesis that dynamically routing instances to specialized contexts unlocks substantial performance gains. On average, NCCE achieves a test accuracy of 74.8\%, outperforming the strongest global baseline POLCA (71.5\%) by an absolute margin of 3.3\%. This superiority holds across varying forms of instance heterogeneity, with NCCE reaching 74.7\% on HoVer, 89.7\% on SCONE, and 60.1\% on HotpotQA, comfortably beating the best-performing baselines on each respective dataset. NCCE also outperforms instance-adaptive baselines Prompt-OIRL (70.5\%) and MoP (68.2\%), which route over a fixed catalog, confirming that catalog co-evolution provides gains beyond routing alone. Overall, the results strongly demonstrate that learning instance-context preferences with a co-evolved catalog is far more effective than either a single global strategy or a fixed-catalog routing approach, even when optimized by state-of-the-art methods.

\subsection{Ablation Study}

Table~\ref{tab:ablation} details our ablation study to evaluate the contribution of each NCCE component:

\begin{itemize}[leftmargin=10pt]

\item \textbf{No routing.}
Applies a single globally optimal strategy from the final evolved catalog to all instances, isolating the contribution of instance-wise routing.


\item \textbf{Random routing.}
Randomly assigns a strategy from the final catalog to each instance, testing whether gains come from learned routing rather than merely expanding the context catalog.


\item \textbf{Cluster-only routing.}
Uses the initial cluster-based context catalog before Context-CF Co-Evolution while keeping the same NCF-based routing mechanism as NCCE, isolating the contribution of context catalog expansion.


\item \textbf{Pointwise loss.}
Replaces the pairwise ranking objective with pointwise regression for preference prediction, evaluating the importance of relative instance-context preference modeling.


\item \textbf{Oracle routing.}
Assigns each instance the ground-truth optimal strategy from the final catalog, providing an upper bound for routing performance.


\end{itemize}

As shown in Table~\ref{tab:ablation}, removing or replacing any core component of NCCE leads to a noticeable performance drop, confirming their respective contributions. First, simply maintaining a larger context catalog without intelligent routing actively harms performance, as seen by \textit{Random routing} yielding the lowest average accuracy (69.2\%). Furthermore, extracting a single best strategy from the evolved pool (\textit{No routing}, 72.0\%) falls short of full NCCE (74.8\%), reinforcing that a global optimum cannot satisfy all instances. Both variants use the final evolved catalog and therefore preserve the context strategies available to full NCCE. No routing applies one strategy to every instance, while Random routing assigns strategies without using the learned instance-context compatibility scores. These settings distinguish the availability of candidate contexts from the way they are selected. In particular, retaining the evolved catalog allows the comparison to focus on the selection mechanism, rather than comparing strategies produced by different context optimization procedures.

The necessity of the co-evolution phase is validated by \textit{Cluster-only routing} (72.4\%), which shows that while initial cluster-level anchors are helpful, iteratively evolving contexts specifically for failure instances is crucial for pushing accuracy higher. Cluster-only routing retains the same NCF-based routing mechanism as full NCCE, but restricts the candidate catalog to the cluster-specific anchors obtained during initialization. It therefore preserves instance-wise selection while removing the subsequent context evolution stage. This comparison complements the preceding routing ablations: the routing ablations change how contexts are selected from the final catalog, whereas Cluster-only routing changes the catalog available to the learned selector. Together, these settings examine the two components of NCCE separately and make the contribution of each component easier to interpret. We also observe that training the NCF model with a pairwise ranking loss slightly outperforms \textit{Pointwise loss} (74.8\% vs. 74.3\%), confirming that relative preference learning is better aligned with the ranking nature of context routing. Finally, \textit{Oracle routing} achieves an impressive 84.3\% average accuracy. This not only proves that our co-evolution mechanism successfully generates a highly capable and diverse context catalog, but also indicates substantial headroom for future advancements in preference modeling.

\begin{table*}[t]
\centering
\caption{Ablation and routing analysis. Cluster-only routing removes context-CF co-evolution, random routing removes the learned NCF router, and oracle routing reports the upper bound of the final context catalog.}
\label{tab:ablation}
\begin{tabular}{lcccc}
\toprule
\textbf{Method} & \textbf{HoVer} & \textbf{SCONE} & \textbf{HotpotQA} & \textbf{Average} \\
\midrule
No routing (global optimal) & 73.5 & 84.7 & 57.8 & 72.0 \\
Random routing              & 72.7 & 81.3 & 53.5 & 69.2 \\
Cluster-only routing        & 73.8 & 89.0 & 54.4 & 72.4 \\
NCCE with pointwise loss    & 73.7 & 89.4 & 59.8 & 74.3 \\
NCCE full          & \underline{74.7} & \underline{89.7} & \underline{60.1} & \underline{74.8} \\
Oracle routing              & \textbf{86.0} & \textbf{95.4} & \textbf{71.6} & \textbf{84.3} \\
\bottomrule
\end{tabular}
\end{table*}

\subsection{Co-Evolution Effect}

\begin{figure*}[t]
    \centering
    \includegraphics[width=\linewidth]{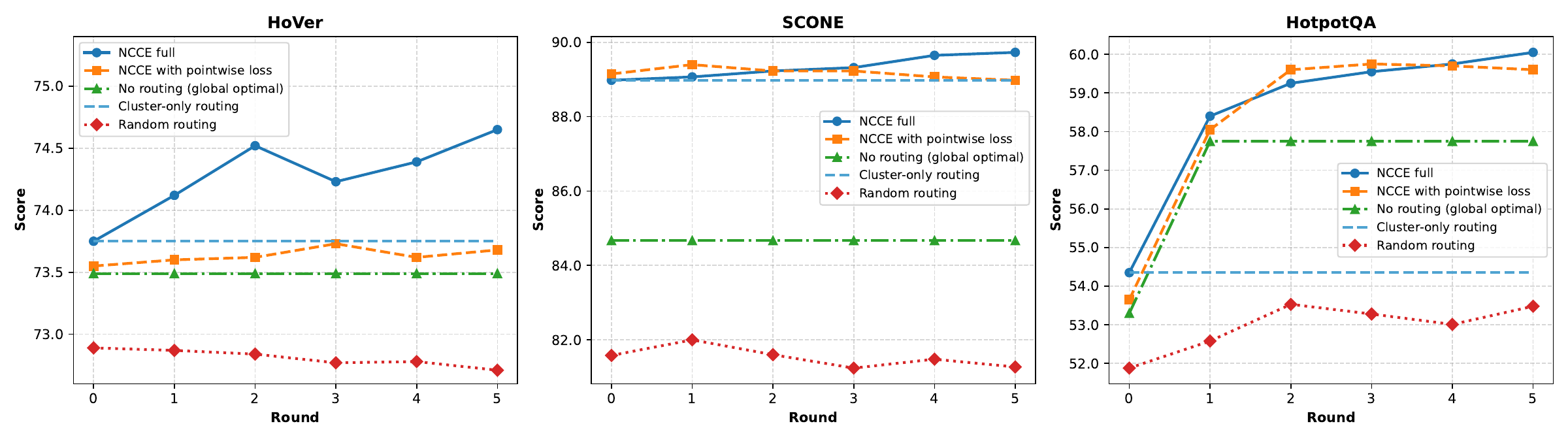}
    \caption{Performance evolution across iterative rounds. The curves track the task scores of NCCE and its ablation variants over six rounds (Round 0--5) on HoVer, SCONE, and HotpotQA. While the full NCCE framework exhibits a sustained upward trajectory, ablated variants either plateau early or show limited gains, demonstrating the effectiveness of the co-evolution process.}
    \label{fig:evolution_curves}
\end{figure*}

To analyze the dynamics of the Context-CF Co-Evolution, we track the task scores as the context catalog expands and the preference model refines. These performance are illustrated in Figure~\ref{fig:evolution_curves}.

\textbf{Effectiveness of Context-CF Co-Evolution.} The evolution curves across all datasets demonstrate a consistent upward trajectory for the full NCCE model over five rounds. This proves that leveraging the preference model to target failure instances successfully generates specialized contexts that continuously enhance overall accuracy. Furthermore, compared to the pointwise loss variant, which fluctuates in later rounds, NCCE with pairwise ranking maintains a highly stable learning curve, confirming its robustness in integrating newly evolved contexts.

\textbf{Necessity of Instance-Wise Routing.} The curves also starkly highlight the limitations of global optimization. While the context catalog expands and improves, the "No routing (global optimal)" or ``random routing'' baselines quickly plateaus. This indicates that merely generating a diverse pool of high-quality contexts offers marginal gains if the system is constrained to a single, globally averaged strategy. The widening performance gap over successive rounds explicitly proves that dynamic routing is essential to unlock the full potential of an evolved context catalog.

\subsection{Complementary experiments}

\paragraph{Cluster Number Effect}



We evaluate NCCE across varying semantic cluster counts (\(K\)) on the HoVer dataset. Increasing \(K\) from 1 (a global strategy) to 4 or 5 steadily improves test accuracy, confirming that diverse anchor contexts provide superior starting points for instance-wise routing. 

However, at \(K=10\), test performance declines despite continued gains on the development set (>78\%). This is consistent with the finite-sample trade-off suggested by Theorem 1: a larger \(K\) reduces the cluster semantic diameter (lowering the Lipschitz penalty in Term~(I)), but fragments available instances per cluster, potentially degrading anchor quality. Identifying a balanced \(K\) (around 4 or 5) therefore matters for generalization.

\paragraph{Data Density}


\begin{figure}[t]
    \centering
    \begin{minipage}{0.48\linewidth}
        \centering
        \includegraphics[width=\linewidth]{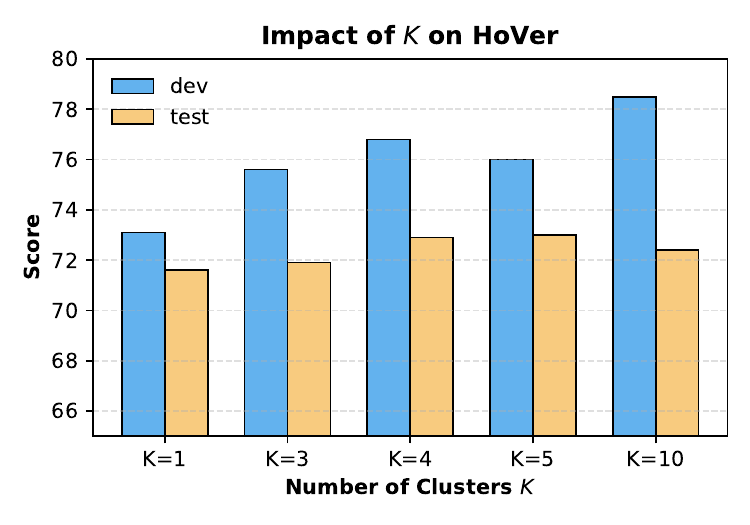}
        \caption{Performance across different cluster numbers.}
        \label{fig:cluster_k}
    \end{minipage}\hfill
    \begin{minipage}{0.48\linewidth}
        \centering
        \includegraphics[width=\linewidth]{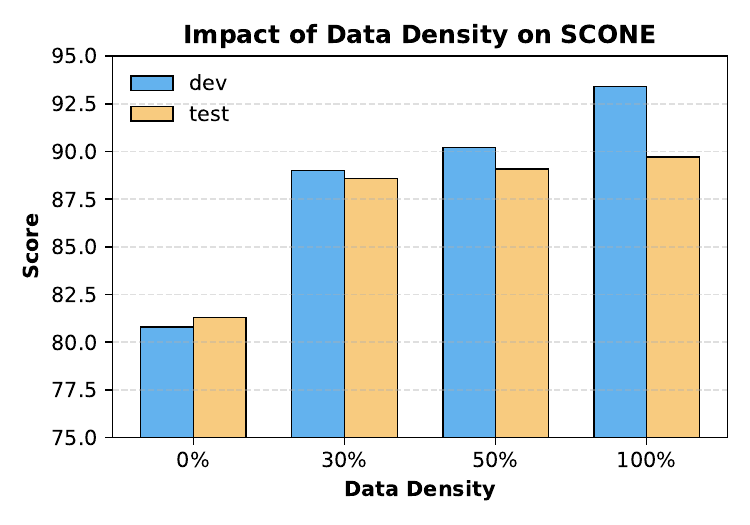}
        \caption{Performance across difference data density in collaborative filtering matrix.}
        \label{fig:data_density}
    \end{minipage}
\end{figure}

To evaluate the robustness of our NCF preference model against interaction sparsity, we vary the proportion of observed instance-context data used to populate the training matrix $\Omega$ on the SCONE dataset. When utilizing 0\% of the interaction data (relying purely on baseline heuristics without collaborative learning), test accuracy sits near 81.3\%. Strikingly, populating the sparse matrix with just 30\% of the available interaction data triggers a massive performance leap, driving test accuracy to roughly 88.6\%. Expanding the data density further to 50\% and 100\% yields diminishing marginal returns on the test set (peaking at 89.7\%), even as development accuracy continues climbing to 93.4\%. This rapid saturation demonstrates that the neural recommender is highly sample-efficient. It successfully learns the latent geometry of instance-context preferences from an extremely sparse matrix, proving that NCCE achieves strong generalization without requiring exhaustive, computationally expensive LLM evaluations across all instance-context pairs.

\paragraph{Routing Context Distribution}

To visualize the routing behavior, we project the instance embeddings into a 2D space using t-SNE, where each point's color represents its assigned context strategy. We quantify the diversity of these assignments using Shannon entropy, defined as $H = -\sum_{j} p_j \log p_j$, where $p_j$ represents the fraction of instances routed to context strategy $j$. Under ``Cluster-only'' routing, the assignments are visibly coarse and dominated by a few rigid spatial regions, reflecting low assignment entropy (e.g., 0.289 on HoVer and 1.308 on SCONE). In contrast, the ``NCCE Full'' framework produces a highly interwoven, heterogeneous distribution of context assignments, driving the entropy significantly higher (1.138 on HoVer and 1.976 on SCONE). Even on HotpotQA, where the entropy remains relatively stable (1.175 vs. 1.147), the spatial mixing of colors in the full model is visually apparent. This confirms that the trained NCF router does not simply default to surface-level semantic neighborhoods. Instead, it successfully captures fine-grained, latent compatibility signals, breaking initial cluster boundaries to route instances based on their specific, nuanced reasoning requirements.

\subsection{Case Study}
\label{sec:case_study}

To complement the aggregate cross-cluster analysis, we present two held-out
SCONE instances that prefer different cluster-specific context strategies.
Under cluster-matched assignment, Instance~A is assigned to $\mathcal{C}_3$ and
receives $p_3$, whereas Instance~B is assigned to $\mathcal{C}_5$ and receives
$p_5$. Here, $p_3$ and $p_5$ are the MIPROv2-generated anchor contexts for
$\mathcal{C}_3$ and $\mathcal{C}_5$, respectively.
Table~\ref{tab:scone-instance-context-examples} summarizes their semantic
differences and reports correctness under all five cluster-specific anchor
contexts.

The two instances exhibit crossed context preferences. Context $p_3$ solves
Instance~A but fails on Instance~B, whereas $p_5$ shows the reverse pattern.
The remaining three anchor contexts fail on both instances. This pattern is
consistent with the contents of the two strategies: $p_3$ explicitly contrasts
valid and invalid category generalizations under negation, while $p_5$ contains
demonstrations centered on existential statements and person-level category
relations. Although these examples are illustrative rather than representative,
they show that context suitability depends jointly on the input instance and
the context strategy. This interaction motivates instance-wise context routing
rather than selecting a single globally optimized strategy.

The complete instances and context strategies are provided in Appendix~\ref{app:case_details}.

\begin{table*}[t]
\centering
\small
\caption{Two illustrative SCONE instances with different context preferences.
Panel (a) summarizes the reasoning patterns of the instances and anchor
contexts. Panel (b) reports correctness under all five cluster-specific anchor
contexts, where $p_j$ denotes the MIPROv2-generated anchor context for training
cluster $\mathcal{C}_j$. A check mark indicates a correct answer.}
\label{tab:scone-instance-context-examples}

\textbf{(a) Characteristics of the instances and contexts}

\begin{tabular}{p{1.5cm}p{6cm}p{5cm}}
\toprule
\textbf{Component} & \textbf{Brief characteristic} & \textbf{Relevant distinction} \\
\midrule
Instance A
& Negation combined with a subtype-to-supertype relation
  (\texttt{pinscher} $\rightarrow$ \texttt{dog}).
& Not receiving a pinscher does not imply not receiving any dog; the correct
  answer is \textsc{No}. \\

Instance B
& Nested negation and existential subtype-to-supertype entailment
  (\texttt{child} $\rightarrow$ \texttt{person}).
& The existence of a walking child entails the existence of a walking person;
  the correct answer is \textsc{Yes}. \\

Context $p_3$
& Its instruction explicitly emphasizes synonyms, category relationships, and
  negation. Its demonstrations contrast valid and invalid category
  generalizations under different polarities.
& Particularly aligned with the negated category-relation pattern in
  Instance A. \\

Context $p_5$
& Its demonstrations emphasize existential statements, person-level category
  relations, and reasoning under negation.
& Particularly aligned with the nested-negation and
  \texttt{child} $\rightarrow$ \texttt{person} pattern in Instance B. \\
\bottomrule
\end{tabular}

\textbf{(b) Correctness under each cluster-specific anchor context}

\begin{tabular}{p{2.0cm}p{6.5cm}ccccc}
\toprule
\textbf{Instance} & \textbf{Reasoning pattern} & $\boldsymbol{p_1}$ & $\boldsymbol{p_2}$ & $\boldsymbol{p_3}$ & $\boldsymbol{p_4}$ & $\boldsymbol{p_5}$ \\
\midrule
A & Negated \texttt{pinscher} $\rightarrow$ \texttt{dog}
  & $\times$ & $\times$ & $\checkmark$ & $\times$ & $\times$ \\
B & Nested-negation \texttt{child} $\rightarrow$ \texttt{person}
  & $\times$ & $\times$ & $\times$ & $\times$ & $\checkmark$ \\
\bottomrule
\end{tabular}
\end{table*}
\section{Related Work}


\begin{figure*}[t]
    \centering
    \includegraphics[width=\linewidth]{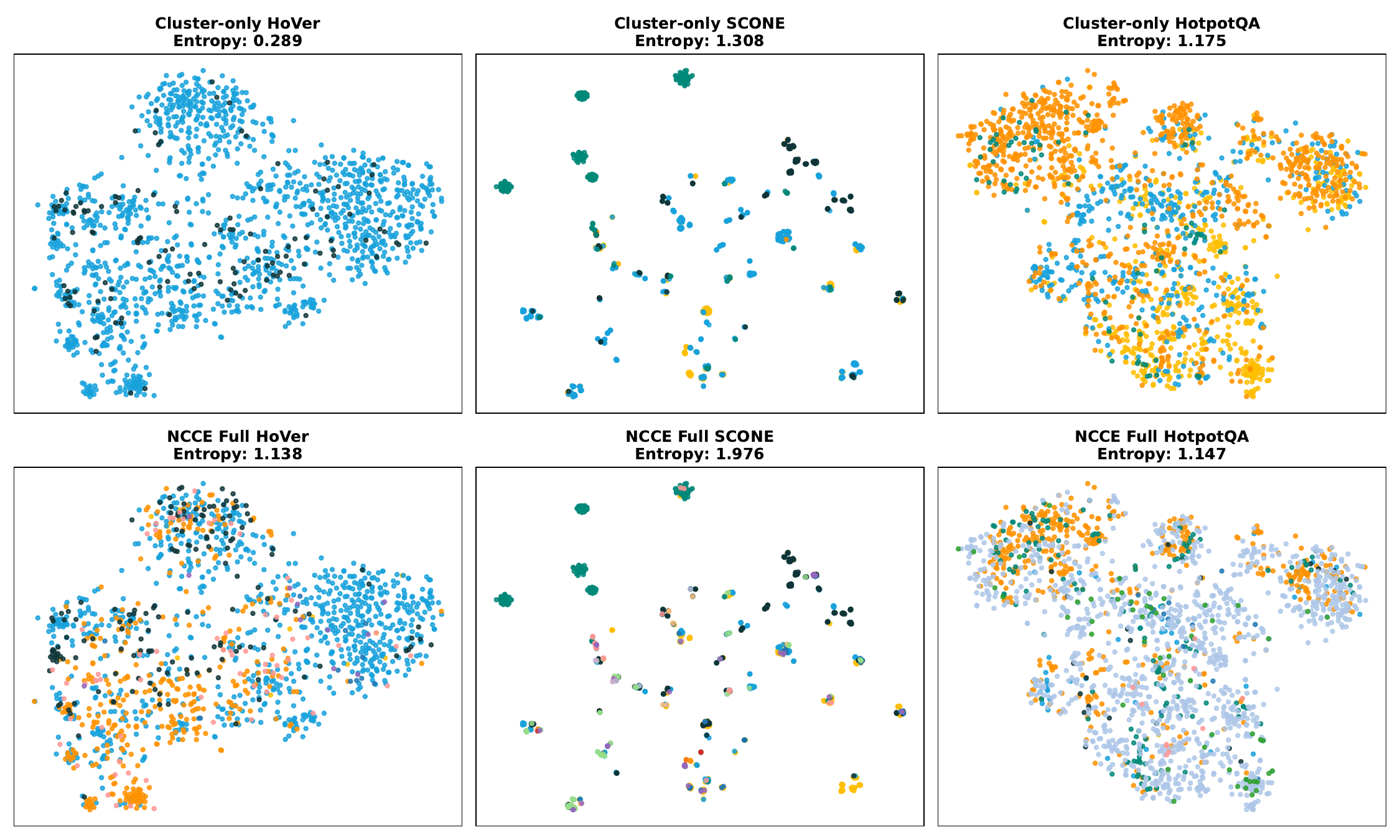}
    
    \caption{t-SNE visualization of context routing assignments. Colors represent different context strategies.}
    
    \label{fig:router_tsne}
\end{figure*}

\subsection{Prompt and Context Engineering}

\textbf{Foundational Prompting and Sensitivity.} Large Language Models (LLMs) are highly sensitive to their input contexts, including instructions, few-shot examples, and reasoning formats. Foundational techniques such as Chain-of-Thought prompting have demonstrated that structuring the input context to include intermediate reasoning steps can significantly elicit complex reasoning capabilities in LLMs \cite{cot}. However, the manual design of these prompts is labor-intensive, motivating the development of automated context optimization techniques.

\textbf{Automated Prompt Optimization.} To systematically discover effective contexts, a substantial body of work treats prompt engineering as an automated search problem. Automatic Prompt Engineer (APE) demonstrates that LLMs can act as human-level prompt engineers by generating and selecting optimal task instructions from a candidate pool \cite{APE}. Similarly, Optimization by PROmpting (OPRO) utilizes LLMs as black-box optimizers to iteratively condition on previous prompts and generate improvements \cite{OPRO}. TextGrad introduces a novel paradigm of automatic differentiation by backpropagating natural language feedback to optimize textual variables \cite{textgrad}. To handle more complex pipelines, MIPROv2 optimizes both instructions and few-shot demonstrations for multi-stage language model programs \cite{miprov2}, while PromptWizard provides a framework specifically designed for task-aware prompt optimization \cite{agarwal2024promptwizardtaskawarepromptoptimization}.

\textbf{Evolutionary and Self-Improving Prompts.} Inspired by biological evolution, several methods maintain a population of prompt candidates and iteratively improve them. EvoPrompt seamlessly connects LLMs with evolutionary algorithms, applying evolutionary operators to yield powerful prompt optimizers \cite{evoprompt}. Pushing this further, Promptbreeder achieves self-referential self-improvement by simultaneously evolving both the prompts and the task-specific mutation operators \cite{fernando2023promptbreederselfreferentialselfimprovementprompt}. Open-source initiatives like OpenEvolve have also extended these evolutionary capabilities to specific domains such as coding agents \cite{openevolve}. Recent advancements highlight the efficiency of this paradigm: GEPA leverages reflective prompt evolution to outperform traditional reinforcement learning \cite{gepa}, and POLCA introduces a stochastic generative optimization framework utilizing LLMs \cite{ren2026polca}.

\textbf{Reflective and Bandit-Based Strategies.} Beyond evolutionary search, recent literature explores self-correction and dynamic exploration. Reflexion equips language agents with verbal reinforcement learning, allowing them to iterate and improve their behavior through generated self-reflections \cite{shinn2023reflexionlanguageagentsverbal}. Other frameworks conceptualize LLMs as in-context bandit reinforcement learners to balance exploration and exploitation during interaction \cite{monea2025llmsincontextbanditreinforcement}. Finally, methods like Rainbow Teaming utilize open-ended generation to produce diverse adversarial prompts, improving model robustness and uncovering failure modes \cite{samvelyan2024rainbowteamingopenendedgeneration}. 

While these methods optimize a single global strategy, ignoring instance-level variation, we propose a Context-CF Co-Evolution mechanism. A lightweight NCF model iteratively generates specialized contexts for failure instances, jointly expanding the catalog and refining instance-context preferences.

\subsection{Instance-Adaptive Prompting and Context Selection}
Beyond globally shared strategies, several works adapt the prompt to individual inputs. One approach retrieves relevant demonstrations per query: EPR trains a dense retriever to find similar few-shot examples \cite{rubin2022learning}, policy-gradient methods select informative exemplars for mathematical reasoning \cite{lu2022dynamic}, and CEIL picks compositionally diverse example sets \cite{ye2023compositional}. Others adapt the instruction or reasoning format directly: TEMPERA uses reinforcement learning to adjust discrete prompt tokens per input \cite{zhang2022tempera}, and instance-adaptive zero-shot CoT tailors the reasoning scaffold to the specific query \cite{yuan2024instance}. A third family maintains a catalog of specialized prompts and routes each input to an appropriate entry. Mixture-of-Prompts (MoP) builds expert prompts from instructions and demonstrations and assigns each query to the nearest cluster centroid \cite{wang2024one}. ConstitutionalExperts groups inputs into semantic clusters and refines principle-based experts using failure feedback \cite{petridis2024constitutionalexperts}. Prompt-OIRL trains a query-dependent reward model from offline execution outcomes to select or request new candidates \cite{sun2024query}, and Automatic Prompt Selection trains an evaluator to rank prompts from a database of clustered training examples \cite{do2024automatic}.

While these methods adapt context selection to individual queries, their catalogs are constructed independently of any preference signal. NCCE co-evolves the two by generating new contexts for failure instances and updating the routing model from their outcomes.


\subsection{Recommendation and Collaborative Filtering}

Recommender systems have a rich history of learning latent preference structures from sparse user-item interaction signals. Early collaborative filtering (CF) approaches relied heavily on neighborhood-based methods, particularly item-to-item similarity algorithms, which proved highly scalable and effective for commercial applications \cite{sarwar2001item, linden2003amazon}. This paradigm evolved significantly with the popularization of Matrix Factorization (MF), which projects users and items into a shared latent space to predict interactions based on inner products \cite{koren2009matrix}. To address severe data sparsity and incorporate arbitrary side features, Factorization Machines (FM) were later introduced as a robust framework for context-aware recommendation \cite{rendle2010factorization}. Crucially, for scenarios driven by implicit feedback, Bayesian Personalized Ranking (BPR) established that optimizing for relative preferences via a pairwise ranking loss yields vastly superior performance compared to absolute pointwise score prediction \cite{rendle2012bpr}.

The integration of deep learning further transformed recommendation architectures by enabling the capture of complex, non-linear user-item relationships. Early neural adaptations successfully applied autoencoders to reconstruct collaborative filtering inputs \cite{sedhain2015autorec}. Subsequently, the industry shifted toward massive deep neural networks capable of multi-stage candidate generation and ranking \cite{covington2016deep}, as well as Wide \& Deep architectures that dynamically balance the memorization of shallow models with the generalization capabilities of deep networks \cite{cheng2016wide}. Most notably, Neural Collaborative Filtering (NCF) generalized traditional MF by replacing the static inner product with a learnable multi-layer perceptron, significantly enhancing the model's expressive power for user-item matching \cite{he2017neural,qin2022rankflow,zhu2024m,qin2023learning}.

Despite these successes, traditional ID-based CF inherently struggles with the cold-start problem, lacking the ability to generalize to unseen users or items. To overcome this limitation, inductive matrix completion methods leverage underlying semantic features (e.g., via graph neural networks) rather than fixed identity embeddings, allowing models to infer preferences for out-of-vocabulary entities \cite{zhang2019inductive}. 

And recently, there emerge lots of works on LLM for recommendation, leveraging LLM's open-world semantic knowledge with traditional CF signals~\cite{zhu2024lifelongpersonalizedlowrankadaptation,shan2025full}.

Our framework is deeply inspired by this technological lineage. By conceptualizing input instances as users and context strategies as items, we frame LLM context engineering as an inductive recommendation problem \cite{zhang2019inductive}. We utilize NCF \cite{he2017neural} trained with a pairwise ranking objective \cite{rendle2012bpr} to perform dynamic, instance-wise context routing, effectively bridging mature CF principles with the frontier of automated prompt optimization.
\section{Limitations}
\label{app:limitation}

While NCCE demonstrates substantial improvements in instance-wise context routing, we acknowledge a few boundaries of our current study that present natural avenues for future research.

First, the Context-CF Co-Evolution phase relies on an LLM reflector to iteratively generate and evaluate new contexts. While this is a one-time, offline training cost and inference remains highly efficient, scaling the co-evolution process to massive datasets incurs unavoidable API latency and computational overhead. Future work could explore using smaller, locally hosted models for the reflection step to reduce dependency on proprietary APIs during training.

Second, our experiments primarily validate the framework using a single, highly capable base LLM. Although the NCF routing mechanism is inherently model-agnostic, exploring the cross-model transferability of the learned context catalog—for instance, evaluating whether contexts evolved using a frontier model can be effectively routed to smaller, open-weight models—remains an interesting open question.

\section{Conclusion}

In this paper, we introduced Neural Collaborative Context Engineering (NCCE), reframing automated context optimization from searching for a single global prompt to a dynamic, instance-wise recommendation problem. NCCE achieves this through a novel Context-CF Co-Evolution mechanism—a synergistic feedback loop between a lightweight NCF model and an LLM reflector that iteratively expands a catalog of specialized contexts. At inference, the NCF model efficiently routes each query to its optimal strategy. Experiments across reasoning benchmarks (HoVer, SCONE, HotpotQA) demonstrate that NCCE significantly outperforms both global optimization baselines and instance-adaptive methods that route over a fixed catalog. Ultimately, our analyses confirm that co-evolving the context catalog alongside preference learning unlocks substantial performance gains, providing a scalable pathway toward highly adaptive LLM inference.

\bibliographystyle{ACM-Reference-Format}
\bibliography{custom, texts/reference}

@String{Computing = "Computing" }

@String{Computer = "{IEEE} Computer" }

@ArtifactSoftware{R,
    title = {R: A Language and Environment for Statistical Computing},
    author = {{R Core Team}},
    organization = {R Foundation for Statistical Computing},
    address = {Vienna, Austria},
    year = {2019},
    url = {https://www.R-project.org/},
}

@misc{APE,
      title={Large Language Models Are Human-Level Prompt Engineers}, 
      author={Yongchao Zhou and Andrei Ioan Muresanu and Ziwen Han and Keiran Paster and Silviu Pitis and Harris Chan and Jimmy Ba},
      year={2023},
      eprint={2211.01910},
      archivePrefix={arXiv},
      primaryClass={cs.LG},
      url={https://arxiv.org/abs/2211.01910}, 
}

@misc{OPRO,
      title={Large Language Models as Optimizers}, 
      author={Chengrun Yang and Xuezhi Wang and Yifeng Lu and Hanxiao Liu and Quoc V. Le and Denny Zhou and Xinyun Chen},
      year={2024},
      eprint={2309.03409},
      archivePrefix={arXiv},
      primaryClass={cs.LG},
      url={https://arxiv.org/abs/2309.03409}, 
}

@misc{cot,
      title={Chain-of-Thought Prompting Elicits Reasoning in Large Language Models}, 
      author={Jason Wei and Xuezhi Wang and Dale Schuurmans and Maarten Bosma and Brian Ichter and Fei Xia and Ed Chi and Quoc Le and Denny Zhou},
      year={2023},
      eprint={2201.11903},
      archivePrefix={arXiv},
      primaryClass={cs.CL},
      url={https://arxiv.org/abs/2201.11903}, 
}

@misc{textgrad,
      title={TextGrad: Automatic "Differentiation" via Text}, 
      author={Mert Yuksekgonul and Federico Bianchi and Joseph Boen and Sheng Liu and Zhi Huang and Carlos Guestrin and James Zou},
      year={2024},
      eprint={2406.07496},
      archivePrefix={arXiv},
      primaryClass={cs.CL},
      url={https://arxiv.org/abs/2406.07496}, 
}

@misc{evoprompt,
      title={EvoPrompt: Connecting LLMs with Evolutionary Algorithms Yields Powerful Prompt Optimizers}, 
      author={Qingyan Guo and Rui Wang and Junliang Guo and Bei Li and Kaitao Song and Xu Tan and Guoqing Liu and Jiang Bian and Yujiu Yang},
      year={2025},
      eprint={2309.08532},
      archivePrefix={arXiv},
      primaryClass={cs.CL},
      url={https://arxiv.org/abs/2309.08532}, 
}

@misc{agarwal2024promptwizardtaskawarepromptoptimization,
      title={PromptWizard: Task-Aware Prompt Optimization Framework}, 
      author={Eshaan Agarwal and Joykirat Singh and Vivek Dani and Raghav Magazine and Tanuja Ganu and Akshay Nambi},
      year={2024},
      eprint={2405.18369},
      archivePrefix={arXiv},
      primaryClass={cs.CL},
      url={https://arxiv.org/abs/2405.18369}, 
}

@misc{fernando2023promptbreederselfreferentialselfimprovementprompt,
      title={Promptbreeder: Self-Referential Self-Improvement Via Prompt Evolution}, 
      author={Chrisantha Fernando and Dylan Banarse and Henryk Michalewski and Simon Osindero and Tim Rocktäschel},
      year={2023},
      eprint={2309.16797},
      archivePrefix={arXiv},
      primaryClass={cs.CL},
      url={https://arxiv.org/abs/2309.16797}, 
}

@software{openevolve,
  title = {OpenEvolve: an open-source evolutionary coding agent},
  author = {Asankhaya Sharma},
  year = {2025},
  publisher = {GitHub},
  url = {https://github.com/algorithmicsuperintelligence/openevolve}
}

@misc{miprov2,
      title={Optimizing Instructions and Demonstrations for Multi-Stage Language Model Programs}, 
      author={Krista Opsahl-Ong and Michael J Ryan and Josh Purtell and David Broman and Christopher Potts and Matei Zaharia and Omar Khattab},
      year={2024},
      eprint={2406.11695},
      archivePrefix={arXiv},
      primaryClass={cs.CL},
      url={https://arxiv.org/abs/2406.11695}, 
}

@misc{gepa,
      title={GEPA: Reflective Prompt Evolution Can Outperform Reinforcement Learning}, 
      author={Lakshya A Agrawal and Shangyin Tan and Dilara Soylu and Noah Ziems and Rishi Khare and Krista Opsahl-Ong and Arnav Singhvi and Herumb Shandilya and Michael J Ryan and Meng Jiang and Christopher Potts and Koushik Sen and Alexandros G. Dimakis and Ion Stoica and Dan Klein and Matei Zaharia and Omar Khattab},
      year={2026},
      eprint={2507.19457},
      archivePrefix={arXiv},
      primaryClass={cs.CL},
      url={https://arxiv.org/abs/2507.19457}, 
}

@misc{samvelyan2024rainbowteamingopenendedgeneration,
      title={Rainbow Teaming: Open-Ended Generation of Diverse Adversarial Prompts}, 
      author={Mikayel Samvelyan and Sharath Chandra Raparthy and Andrei Lupu and Eric Hambro and Aram H. Markosyan and Manish Bhatt and Yuning Mao and Minqi Jiang and Jack Parker-Holder and Jakob Foerster and Tim Rocktäschel and Roberta Raileanu},
      year={2024},
      eprint={2402.16822},
      archivePrefix={arXiv},
      primaryClass={cs.CL},
      url={https://arxiv.org/abs/2402.16822}, 
}

@misc{shinn2023reflexionlanguageagentsverbal,
      title={Reflexion: Language Agents with Verbal Reinforcement Learning}, 
      author={Noah Shinn and Federico Cassano and Edward Berman and Ashwin Gopinath and Karthik Narasimhan and Shunyu Yao},
      year={2023},
      eprint={2303.11366},
      archivePrefix={arXiv},
      primaryClass={cs.AI},
      url={https://arxiv.org/abs/2303.11366}, 
}

@misc{monea2025llmsincontextbanditreinforcement,
      title={LLMs Are In-Context Bandit Reinforcement Learners}, 
      author={Giovanni Monea and Antoine Bosselut and Kianté Brantley and Yoav Artzi},
      year={2025},
      eprint={2410.05362},
      archivePrefix={arXiv},
      primaryClass={cs.CL},
      url={https://arxiv.org/abs/2410.05362}, 
}

@article{ren2026polca,
  title={POLCA: Stochastic Generative Optimization with LLM},
  author={Ren, Xuanfei and Nie, Allen and Xie, Tengyang and Cheng, Ching-An},
  journal={arXiv preprint arXiv:2603.14769},
  year={2026}
}

@inproceedings{sarwar2001item,
  title={Item-based collaborative filtering recommendation algorithms},
  author={Sarwar, Badrul and Karypis, George and Konstan, Joseph and Riedl, John},
  booktitle={Proceedings of the 10th international conference on World Wide Web},
  pages={285--295},
  year={2001}
}

@article{linden2003amazon,
  title={Amazon. com recommendations: Item-to-item collaborative filtering},
  author={Linden, Greg and Smith, Brent and York, Jeremy},
  journal={IEEE Internet computing},
  volume={7},
  number={1},
  pages={76--80},
  year={2003},
  publisher={IEEE}
}

@article{koren2009matrix,
  title={Matrix factorization techniques for recommender systems},
  author={Koren, Yehuda and Bell, Robert and Volinsky, Chris},
  journal={Computer},
  volume={42},
  number={8},
  pages={30--37},
  year={2009},
  publisher={IEEE}
}

@article{rendle2012bpr,
  title={BPR: Bayesian personalized ranking from implicit feedback},
  author={Rendle, Steffen and Freudenthaler, Christoph and Gantner, Zeno and Schmidt-Thieme, Lars},
  journal={arXiv preprint arXiv:1205.2618},
  year={2012}
}

@inproceedings{rendle2010factorization,
  title={Factorization machines},
  author={Rendle, Steffen},
  booktitle={2010 IEEE International conference on data mining},
  pages={995--1000},
  year={2010},
  organization={IEEE}
}

@inproceedings{he2017neural,
  title={Neural collaborative filtering},
  author={He, Xiangnan and Liao, Lizi and Zhang, Hanwang and Nie, Liqiang and Hu, Xia and Chua, Tat-Seng},
  booktitle={Proceedings of the 26th international conference on world wide web},
  pages={173--182},
  year={2017}
}

@inproceedings{covington2016deep,
  title={Deep neural networks for youtube recommendations},
  author={Covington, Paul and Adams, Jay and Sargin, Emre},
  booktitle={Proceedings of the 10th ACM conference on recommender systems},
  pages={191--198},
  year={2016}
}

@inproceedings{cheng2016wide,
  title={Wide \& deep learning for recommender systems},
  author={Cheng, Heng-Tze and Koc, Levent and Harmsen, Jeremiah and Shaked, Tal and Chandra, Tushar and Aradhye, Hrishi and Anderson, Glen and Corrado, Greg and Chai, Wei and Ispir, Mustafa and others},
  booktitle={Proceedings of the 1st workshop on deep learning for recommender systems},
  pages={7--10},
  year={2016}
}

@article{zhang2019inductive,
  title={Inductive matrix completion based on graph neural networks},
  author={Zhang, Muhan and Chen, Yixin},
  journal={arXiv preprint arXiv:1904.12058},
  year={2019}
}

@inproceedings{sedhain2015autorec,
  title={Autorec: Autoencoders meet collaborative filtering},
  author={Sedhain, Suvash and Menon, Aditya Krishna and Sanner, Scott and Xie, Lexing},
  booktitle={Proceedings of the 24th international conference on World Wide Web},
  pages={111--112},
  year={2015}
}

@inproceedings{jiang2020hover,
  title={HoVer: A dataset for many-hop fact extraction and claim verification},
  author={Jiang, Yichen and Bordia, Shikha and Zhong, Zheng and Dognin, Charles and Singh, Maneesh and Bansal, Mohit},
  booktitle={Findings of the Association for Computational Linguistics: EMNLP 2020},
  pages={3441--3460},
  year={2020}
}

@inproceedings{scone,
  title={Simpler context-dependent logical forms via model projections},
  author={Long, Reginald and Pasupat, Panupong and Liang, Percy},
  booktitle={Proceedings of the 54th Annual Meeting of the Association for Computational Linguistics (Volume 1: Long Papers)},
  pages={1456--1465},
  year={2016}
}

@inproceedings{yang2018hotpotqa,
  title={HotpotQA: A dataset for diverse, explainable multi-hop question answering},
  author={Yang, Zhilin and Qi, Peng and Zhang, Saizheng and Bengio, Yoshua and Cohen, William and Salakhutdinov, Ruslan and Manning, Christopher D},
  booktitle={Proceedings of the 2018 conference on empirical methods in natural language processing},
  pages={2369--2380},
  year={2018}
}

@misc{openai2024gpt4ocard,
      title={GPT-4o System Card}, 
      author={OpenAI and : and Aaron Hurst and Adam Lerer and Adam P. Goucher and etc.},
      year={2024},
      eprint={2410.21276},
      archivePrefix={arXiv},
      primaryClass={cs.CL},
      url={https://arxiv.org/abs/2410.21276}, 
}

@article{achiam2023gpt,
  title={Gpt-4 technical report},
  author={Achiam, Josh and Adler, Steven and Agarwal, Sandhini and Ahmad, Lama and Akkaya, Ilge and Aleman, Florencia Leoni and Almeida, Diogo and Altenschmidt, Janko and Altman, Sam and Anadkat, Shyamal and others},
  journal={arXiv preprint arXiv:2303.08774},
  year={2023}
}

@article{brown2020language,
  title={Language models are few-shot learners},
  author={Brown, Tom and Mann, Benjamin and Ryder, Nick and Subbiah, Melanie and Kaplan, Jared D and Dhariwal, Prafulla and Neelakantan, Arvind and Shyam, Pranav and Sastry, Girish and Askell, Amanda and others},
  journal={Advances in neural information processing systems},
  volume={33},
  pages={1877--1901},
  year={2020}
}

@article{touvron2023llama,
  title={Llama 2: open foundation and fine-tuned chat models. arXiv},
  author={Touvron, Hugo and Martin, Louis and Stone, Kevin and Albert, Peter and Almahairi, Amjad and Babaei, Yasmine and Bashlykov, Nikolay and Batra, Soumya and Bhargava, Prajjwal and Bhosale, Shruti and others},
  journal={arXiv preprint arXiv:2307.09288},
  volume={10},
  year={2023},
  publisher={Retrieved 2023-11-14, from http://arxiv. org/abs/2307.09288}
}

@inproceedings{zhao2021calibrate,
  title={Calibrate before use: Improving few-shot performance of language models},
  author={Zhao, Zihao and Wallace, Eric and Feng, Shi and Klein, Dan and Singh, Sameer},
  booktitle={International conference on machine learning},
  pages={12697--12706},
  year={2021},
  organization={Pmlr}
}

@article{zhao2023survey,
  title={A survey of large language models},
  author={Zhao, Wayne Xin and Zhou, Kun and Li, Junyi and Tang, Tianyi and Wang, Xiaolei and Hou, Yupeng and Min, Yingqian and Zhang, Beichen and Zhang, Junjie and Dong, Zican and others},
  journal={arXiv preprint arXiv:2303.18223},
  volume={1},
  number={2},
  pages={1--124},
  year={2023}
}

@inproceedings{khattab2023dspy,
  title={DSPy: compiling declarative language model calls into state-of-the-art pipelines},
  author={Khattab, Omar and Singhvi, Arnav and Maheshwari, Paridhi and Zhang, Zhiyuan and Santhanam, Keshav and Haq, Saiful and Sharma, Ashutosh and Joshi, Thomas T and Moazam, Hanna and Miller, Heather and others},
  booktitle={The Twelfth International Conference on Learning Representations},
  year={2023}
}

@article{wang2020minilm,
  title={Minilm: Deep self-attention distillation for task-agnostic compression of pre-trained transformers},
  author={Wang, Wenhui and Wei, Furu and Dong, Li and Bao, Hangbo and Yang, Nan and Zhou, Ming},
  journal={Advances in neural information processing systems},
  volume={33},
  pages={5776--5788},
  year={2020}
}

@article{wang2024openr,
  title={Openr: An open source framework for advanced reasoning with large language models},
  author={Wang, Jun and Fang, Meng and Wan, Ziyu and Wen, Muning and Zhu, Jiachen and Liu, Anjie and Gong, Ziqin and Song, Yan and Chen, Lei and Ni, Lionel M and others},
  journal={arXiv preprint arXiv:2410.09671},
  year={2024}
}

@inproceedings{qin2022rankflow,
  title={Rankflow: Joint optimization of multi-stage cascade ranking systems as flows},
  author={Qin, Jiarui and Zhu, Jiachen and Chen, Bo and Liu, Zhirong and Liu, Weiwen and Tang, Ruiming and Zhang, Rui and Yu, Yong and Zhang, Weinan},
  booktitle={Proceedings of the 45th International ACM SIGIR Conference on Research and Development in Information Retrieval},
  pages={814--824},
  year={2022}
}

@misc{zhu2024lifelongpersonalizedlowrankadaptation,
      title={Lifelong Personalized Low-Rank Adaptation of Large Language Models for Recommendation}, 
      author={Jiachen Zhu and Jianghao Lin and Xinyi Dai and Bo Chen and Rong Shan and Jieming Zhu and Ruiming Tang and Yong Yu and Weinan Zhang},
      year={2024},
      eprint={2408.03533},
      archivePrefix={arXiv},
      primaryClass={cs.IR},
      url={https://arxiv.org/abs/2408.03533}, 
}

@inproceedings{zhu2024m,
  title={M-scan: A multi-scenario causal-driven adaptive network for recommendation},
  author={Zhu, Jiachen and Wang, Yichao and Lin, Jianghao and Qin, Jiarui and Tang, Ruiming and Zhang, Weinan and Yu, Yong},
  booktitle={Proceedings of the ACM Web Conference 2024},
  pages={3844--3853},
  year={2024}
}

@inproceedings{qin2023learning,
  title={Learning to distinguish multi-user coupling behaviors for TV recommendation},
  author={Qin, Jiarui and Zhu, Jiachen and Liu, Yankai and Gao, Junchao and Ying, Jianjie and Liu, Chaoxiong and Wang, Ding and Feng, Junlan and Deng, Chao and Wang, Xiaozheng and others},
  booktitle={Proceedings of the sixteenth ACM international conference on web search and data mining},
  pages={204--212},
  year={2023}
}

@article{shan2025full,
  title={Full-Stack Optimized Large Language Models for Lifelong Sequential Behavior Comprehension in Recommendation},
  author={Shan, Rong and Zhu, Jiachen and Lin, Jianghao and Zhu, Chenxu and Chen, Bo and Tang, Ruiming and Yu, Yong and Zhang, Weinan},
  journal={ACM Transactions on Recommender Systems},
  volume={4},
  number={2},
  pages={1--33},
  year={2025},
  publisher={ACM New York, NY}
}

@article{wang2024one,
  title={One prompt is not enough: Automated construction of a mixture-of-expert prompts},
  author={Wang, Ruochen and An, Sohyun and Cheng, Minhao and Zhou, Tianyi and Hwang, Sung Ju and Hsieh, Cho-Jui},
  journal={arXiv preprint arXiv:2407.00256},
  year={2024}
}

@inproceedings{sun2024query,
  title={Query-dependent prompt evaluation and optimization with offline inverse rl},
  author={Sun, Hao and H{\"u}y{\"u}k, Alihan and van der Schaar, Mihaela},
  booktitle={International Conference on Learning Representations},
  volume={2024},
  pages={42215--42270},
  year={2024}
}

@inproceedings{petridis2024constitutionalexperts,
  title={ConstitutionalExperts: Training a mixture of principle-based prompts},
  author={Petridis, Savvas and Wedin, Ben and Yuan, Ann and Wexler, James and Thain, Nithum},
  booktitle={Proceedings of the 62nd Annual Meeting of the Association for Computational Linguistics (Volume 2: Short Papers)},
  pages={574--582},
  year={2024}
}

@article{do2024automatic,
  title={Automatic prompt selection for large language models},
  author={Do, Viet-Tung and Hoang, Van-Khanh and Nguyen, Duy-Hung and Sabahi, Shahab and Yang, Jeff and Hotta, Hajime and Nguyen, Minh-Tien and Le, Hung},
  journal={arXiv preprint arXiv:2404.02717},
  year={2024}
}

@article{zhang2022tempera,
  title={Tempera: Test-time prompting via reinforcement learning},
  author={Zhang, Tianjun and Wang, Xuezhi and Zhou, Denny and Schuurmans, Dale and Gonzalez, Joseph E},
  journal={arXiv preprint arXiv:2211.11890},
  year={2022}
}

@article{yuan2024instance,
  title={Instance-adaptive zero-shot chain-of-thought prompting},
  author={Yuan, Xiaosong and Shen, Chen and Yan, Shaotian and Zhang, Xiaofeng and Xie, Liang and Wang, Wenxiao and Guan, Renchu and Wang, Ying and Ye, Jieping},
  journal={Advances in Neural Information Processing Systems},
  volume={37},
  pages={125469--125486},
  year={2024}
}

@inproceedings{rubin2022learning,
  title={Learning to retrieve prompts for in-context learning},
  author={Rubin, Ohad and Herzig, Jonathan and Berant, Jonathan},
  booktitle={Proceedings of the 2022 conference of the North American chapter of the association for computational linguistics: human language technologies},
  pages={2655--2671},
  year={2022}
}

@article{lu2022dynamic,
  title={Dynamic prompt learning via policy gradient for semi-structured mathematical reasoning},
  author={Lu, Pan and Qiu, Liang and Chang, Kai-Wei and Wu, Ying Nian and Zhu, Song-Chun and Rajpurohit, Tanmay and Clark, Peter and Kalyan, Ashwin},
  journal={arXiv preprint arXiv:2209.14610},
  year={2022}
}

@inproceedings{ye2023compositional,
  title={Compositional exemplars for in-context learning},
  author={Ye, Jiacheng and Wu, Zhiyong and Feng, Jiangtao and Yu, Tao and Kong, Lingpeng},
  booktitle={International Conference on Machine Learning},
  pages={39818--39833},
  year={2023},
  organization={PMLR}
}

@inproceedings{she2023scone,
  title={ScoNe: Benchmarking negation reasoning in language models with fine-tuning and in-context learning},
  author={She, Jingyuan S and Potts, Christopher and Bowman, Samuel R and Geiger, Atticus},
  booktitle={Proceedings of the 61st Annual Meeting of the Association for Computational Linguistics (Volume 2: Short Papers)},
  pages={1803--1821},
  year={2023}
}

@article{bartlett2002rademacher,
  title={Rademacher and gaussian complexities: Risk bounds and structural results},
  author={Bartlett, Peter L and Mendelson, Shahar},
  journal={Journal of machine learning research},
  volume={3},
  number={Nov},
  pages={463--482},
  year={2002}
}

\appendix

\section{Notation}
\label{app:notation}

Table~\ref{tab:notation} summarizes the notation used in the main text and appendices.

\begingroup
\renewcommand{\arraystretch}{1}
\begin{table*}[h]
\centering
\begin{tabularx}{\textwidth}{>{\raggedright\arraybackslash}p{0.22\textwidth}
                              >{\raggedright\arraybackslash}X}
\toprule
\textbf{Symbol} & \textbf{Description} \\
\midrule
\multicolumn{2}{l}{\textbf{Problem formulation and representations}}\\
\addlinespace[2pt]
\(X=\{x_1,\ldots,x_N\}\) & Set of input or training instances. \\
\(x_i\) & The \(i\)-th input instance. \\
\(N\) & Number of input instances. \\
\(\mathcal{X}\) & Input space. \\
\(\mathcal{D}\) & Distribution over the input space \(\mathcal{X}\). \\
\(P=\{p_1,\ldots,p_M\}\) & Catalog of candidate context strategies. \\
\(p_j\) & The \(j\)-th context strategy; in the routing-entropy analysis, the same symbol is reused for the fraction of instances routed to strategy \(j\). \\
\(M\) & Number of context strategies in the catalog. \\
\(c_j^{\mathrm{inst}}\) & Task-instruction component of context strategy \(p_j\). \\
\(c_j^{\mathrm{demo}}\) & Few-shot-demonstration component of context strategy \(p_j\). \\
\(c_j^{\mathrm{reason}}\) & Reasoning-format component of context strategy \(p_j\). \\
\(c_j^{\mathrm{out}}\) & Output-constraint component of context strategy \(p_j\). \\
\(R(x_i,p_j)\) & Evaluation function returning the task accuracy or reward obtained by applying strategy \(p_j\) to instance \(x_i\). \\
\(r_{ij}\) & Observed task accuracy of strategy \(p_j\) on instance \(x_i\). \\
\(r(x,p)\) & True reward of applying context \(p\) to instance \(x\). \\
\(p^*\) & Globally optimal context strategy shared across all instances. \\
\(p_i^*\) & Optimal context strategy for instance \(x_i\). \\
\(\Omega\) & Sparse set of observed instance--context interactions. \\
\(\hat r_{ij}\) & Predicted compatibility score for instance \(x_i\) and context strategy \(p_j\). \\
\(\phi(\cdot)\) & Frozen text encoder for input instances. \\
\(\psi(\cdot)\) & Frozen text encoder for context strategies. \\
\(e_i=\phi(x_i)\) & Semantic representation of instance \(x_i\). \\
\(h_j=\psi(p_j)\) & Semantic representation of context strategy \(p_j\). \\
\(W_x,\,W_p\) & Projection matrices that map instance and context representations into a shared latent space. \\
\(u_i=W_xe_i\) & Projected latent representation of instance \(x_i\). \\
\(v_j=W_ph_j\) & Projected latent representation of context strategy \(p_j\). \\
\(z_{ij}\) & Interaction vector \([u_i;v_j;u_i\odot v_j;|u_i-v_j|]\). \\
\(f_\theta\) & Neural collaborative filtering compatibility model. \\
\(\theta\) & Learnable parameters of the NCF compatibility model. \\
\(\mathrm{MLP}_\theta\) & Multi-layer perceptron used to compute compatibility scores. \\
\(\sigma\) & Logistic sigmoid function, \(\sigma(a)=(1+e^{-a})^{-1}\). \\

\bottomrule
\end{tabularx}
\caption{Notation used throughout the paper.}
\label{tab:notation}
\end{table*}

\begin{table*}[h]
\ContinuedFloat
\centering
\begin{tabularx}{\textwidth}{>{\raggedright\arraybackslash}p{0.22\textwidth}
                              >{\raggedright\arraybackslash}X}
\toprule
\textbf{Symbol} & \textbf{Description} \\
\midrule

\multicolumn{2}{l}{\textbf{Preference learning and catalog initialization}}\\
\addlinespace[2pt]
\(\mathcal{D}_{\mathrm{pair}}\) & Set of observed preference triples \((i,j,k)\), where \(p_j\) outperforms \(p_k\) on \(x_i\). \\
\(\mathcal{L}_{\mathrm{rank}}\) & Pairwise ranking loss. \\
\(\mathcal{C}_k\) & The \(k\)-th semantic cluster of input instances. \\
\(K\) & Number of semantic clusters. \\
\(\mathcal{A}\) & Replaceable warm-up optimizer used to generate cluster-specific contexts. \\
\(P_k^0\) & Set of specialized anchor contexts generated for cluster \(\mathcal{C}_k\). \\
\(P_0\) & Initial context catalog, formed by the union of the cluster-specific anchor contexts. \\
\(K^\star\) & Dataset-dependent optimal number of clusters. \\

\addlinespace[4pt]
\multicolumn{2}{l}{\textbf{Context--CF co-evolution}}\\
\addlinespace[2pt]
\(t\) & Co-evolution iteration index. \\
\(T\) & Number of co-evolution iterations. \\
\(P_t\) & Context catalog at iteration \(t\). \\
\(\Omega_t\) & Observed interaction set at iteration \(t\). \\
\(\mathcal{F}_t\) & Set of failure instances not solved by any context in \(P_t\). \\
\(\mathcal{B}_t\) & Batch sampled from the failure set \(\mathcal{F}_t\). \\
\(m\) & Number of failure instances in \(\mathcal{B}_t\); also used as the router-training minibatch size in the experimental appendix. \\
\(k\) & Number of sampled latent seeds in context evolution; elsewhere used as a cluster index or as the lower-ranked context index in a preference triple. \\
\(\mathcal{J}(h;\mathcal{B}_t)\) & Mean NCF-predicted suitability of latent context embedding \(h\) for failure batch \(\mathcal{B}_t\). \\
\(s_\theta(h,x_i)\) & NCF-predicted suitability between latent context embedding \(h\) and instance \(x_i\). \\
\(h^{(\tau)}\) & Latent context embedding after gradient-ascent step \(\tau\). \\
\(\tau\) & Latent gradient-ascent step index; in the router-training appendix, the same symbol denotes the temperature hyperparameter. \\
\(\eta\) & Learning rate used for latent gradient ascent and router training. \\
\(G\) & Number of gradient-ascent steps. \\
\(\tilde h_\ell\) & Optimized latent target embedding for seed \(\ell\). \\
\(\ell\) & Latent-seed index. \\
\(a_\ell\) & Catalog index of the \(\ell\)-th sampled seed context. \\
\(d(p_j)\) & Average Euclidean distance from context \(p_j\) to the optimized latent target embeddings. \\
\(p^{\mathrm{pot}}\) & Existing context selected by minimum average distance to the optimized latent targets. \\
\(\mathcal{M}\) & LLM reflector used to revise the selected context. \\
\(p_{\mathrm{new}}\) & Newly evolved context strategy produced by the LLM reflector. \\
\(\Delta P_t\) & New context strategies generated at iteration \(t\). \\
\(\Delta\Omega_t\) & New observed interactions obtained at iteration \(t\). \\

\bottomrule
\end{tabularx}
\caption{Notation used throughout the paper (continued).}
\end{table*}

\begin{table*}[h]
\ContinuedFloat
\centering
\begin{tabularx}{\textwidth}{>{\raggedright\arraybackslash}p{0.22\textwidth}
                              >{\raggedright\arraybackslash}X}
\toprule
\textbf{Symbol} & \textbf{Description} \\
\midrule

\multicolumn{2}{l}{\textbf{Theoretical analysis}}\\
\addlinespace[2pt]
\(\hat f_\theta:\mathcal{X}\to P\) & Learned instance-wise context router. \\
\(\Delta(x)\) & Instance-wise regret of the learned router. \\
\(L\) & Lipschitz constant in the cluster preference assumption. \\
\(x',\,p'\) & A second instance and context, respectively, used in the cluster Lipschitz preference assumption. \\
\(\delta\) & Confidence parameter; the PAC-style bound holds with probability at least \(1-\delta\). \\
\(n\) & Number of independent inputs drawn i.i.d.\ from \(\mathcal{D}\) used in the generalization bound; a single input may yield multiple pairwise comparisons, but only the number of distinct inputs counts. \\
\(\varepsilon\) & Regret threshold in the tail-probability bound. \\
\(\alpha\) & Population optimality gap of the cluster anchor: \(\alpha_k=\max_{p\in P}\mathbb{E}_{x|\mathcal{C}_k}[r(x,p)-r(x,p_k^0)]\), \(\alpha=\max_k\alpha_k\). \\
\(\rho_K\) & Embedding diameter of the clusters (satisfies \(\rho_K=O(K^{-1/d})\) under regularity conditions on the instance distribution). \\
\(d\) & Intrinsic dimension of the instance-embedding space \(\phi(\mathcal{X})\). \\
\(\mathcal{F}\) & Hypothesis class of NCF-based routers. \\
\(\mathcal{H}\) & Pre-fixed induced loss class consisting of functions \(x\mapsto\ell_h(x)\in[0,1]\) parameterized by the catalog, anchor map, and scorer; used in the anchor-violation Rademacher analysis. \\
\(\ell(x)\) & Anchor-violation loss: \(\max_{q\in P}[r(x,a(x))-r(x,q)]_+\,\mathbf{1}\{s_\theta(x,q)\ge s_\theta(x,a(x))\}\); measures reward lost when the router ranks a suboptimal context above the anchor. \\
\(\mathfrak{R}_n(\mathcal{H})\) & Rademacher complexity of the induced loss class \(\mathcal{H}\). \\
\(\widehat{\mathcal{R}}_n\) & Empirical mean of the anchor-violation loss over \(n\) independent inputs: \(\widehat{\mathcal{R}}_n=\frac{1}{n}\sum_{i=1}^n\ell(x_i)\). \\
\(f^{\mathrm{cluster}}\) & Reference router that maps an instance to its cluster's anchor context. \\
\(p_k^0\) & Anchor context associated with cluster \(\mathcal{C}_k\). \\
\(a(x)\) & Cluster anchor for instance \(x\): \(a(x)=p_k^0\) when \(x\in\mathcal{C}_k\). \\
\(\Delta^{\mathrm{cluster}}(x)\) & Regret of the cluster-anchor reference router. \\

\addlinespace[4pt]
\multicolumn{2}{l}{\textbf{Experimental analysis and hyperparameters}}\\
\addlinespace[2pt]
\(H\) & Shannon entropy of the context-routing assignment distribution. \\
\(p\text{-value}\) & Significance-test \(p\)-value reported in the main-results table. \\
\(\lambda\) & \(L_2\)-regularization (weight-decay) coefficient used to train the NCF router. \\

\addlinespace[4pt]
\multicolumn{2}{l}{\textbf{Operators and indexing conventions}}\\
\addlinespace[2pt]
\(i\) & Instance index. \\
\(j\) & Context-strategy index. \\
\(\odot\) & Element-wise multiplication. \\
\(\nabla_h\) & Gradient with respect to latent context embedding \(h\). \\
\(\lVert\cdot\rVert_2\) & Euclidean norm. \\
\(\operatorname{Normalize}(\cdot)\) & \(\ell_2\)-normalization operator. \\
\(\mathbb{E}[\cdot]\) & Expectation. \\
\(\Pr[\cdot]\) & Probability. \\

\bottomrule
\end{tabularx}
\caption{Notation used throughout the paper (continued).}
\end{table*}
\endgroup

\clearpage
\section{Instances and Context Strategies for the Case Study}
\label{app:case_details}

This appendix reproduces the complete SCONE instances and context strategies used in the case study in Section~\ref{sec:case_study}.
\medskip

\noindent
\begin{minipage}{\linewidth}
\begin{lstlisting}[breaklines=true,breakatwhitespace=false,columns=fullflexible,caption={SCONE Instance A}]
[Dataset: SCONE]
Context: The girl will not get a stuffed pinscher as a gift, but not because she failed the exam.
Question: The girl will not get a stuffed dog as a gift, but not because she failed the exam.
Gold Answer: No
Assigned Cluster: C_3
Selected Context Strategy: p_3
\end{lstlisting}
\end{minipage}

\noindent
\begin{minipage}{\linewidth}
\begin{lstlisting}[breaklines=true,breakatwhitespace=false,columns=fullflexible,caption={SCONE Instance B}]
[Dataset: SCONE]
Context: It is not true that there is not a single child walking in the city.
Question: It is not true that there is not a single person walking in the city.
Gold Answer: Yes
Assigned Cluster: C_5
Selected Context Strategy: p_5
\end{lstlisting}
\end{minipage}

\begin{lstlisting}[breaklines=true,breakatwhitespace=false,columns=fullflexible,caption={SCONE Context Strategy $p_3$}]
[Module: generate_answer]
{
  "predict": {
    "demos": [
      {
        "augmented": true,
        "context": "It is not the case that the girl will not get a stuffed beagle as a gift.",
        "question": "It is not the case that the girl will not get a stuffed dog as a gift.",
        "reasoning": "The context states that it is not the case that the girl will not get a stuffed beagle as a gift, which implies that she will get a stuffed beagle. A beagle is a type of dog, so if she is getting a stuffed beagle, she is indeed getting a stuffed dog. Therefore, the statement that it is not the case that the girl will not get a stuffed dog as a gift is true.",
        "answer": "Yes"
      },
      {
        "augmented": true,
        "context": "The person did not like mangos and did not like meat.",
        "question": "The person did not like fruit and did not like meat.",
        "reasoning": "The context states that the person did not like mangos, which is a specific type of fruit, but it does not explicitly state that they dislike all fruit. Therefore, while it is clear they do not like mangos, this does not necessarily mean they dislike all fruit. The statement about not liking meat is consistent with both the context and the question.",
        "answer": "No"
      },
      {
        "augmented": true,
        "context": "the man owns a pug and he does not drink coffee.",
        "question": "the man owns a dog and he does not drink coffee.",
        "reasoning": "The context states that the man owns a pug. A pug is a breed of dog, so if the man owns a pug, he owns a dog. Additionally, the context clearly mentions that the man does not drink coffee. Therefore, the statement in the question that the man owns a dog and does not drink coffee is consistent with the context provided.",
        "answer": "Yes"
      }
    ],
    "signature": {
      "instructions": "Analyze the given context and question to determine logical consistency. Consider synonyms, category relationships, and negations. Describe your logical reasoning step by step, and conclude with \"Yes\" or \"No,\" indicating whether the question is supported by the context. Ensure that your explanation clarifies any semantic transformations and justifies your answer clearly.",
      "fields": [
        {
          "prefix": "Context:",
          "description": "${context}"
        },
        {
          "prefix": "Question:",
          "description": "${question}"
        },
        {
          "prefix": "Reasoning: Let's think step by step in order to",
          "description": "${reasoning}"
        },
        {
          "prefix": "Answer:",
          "description": "Yes or No"
        }
      ]
    }
  }
}
\end{lstlisting}

\begin{lstlisting}[breaklines=true,breakatwhitespace=false,columns=fullflexible,caption={SCONE Context Strategy $p_5$}]
[Module: generate_answer]
{
  "predict": {
    "demos": [
      {
        "augmented": true,
        "context": "There is a single queen walking in the city that does not sleep and does not rest.",
        "question": "There is a single person walking in the city that does not sleep and does not rest.",
        "reasoning": "The statement in the context refers to a \"single queen\" walking in the city who does not sleep or rest. The question asks whether there is a \"single person\" with the same characteristics. Since a queen is a type of person, the context implies that there is indeed a person who does not sleep or rest, matching the description in the question.",
        "answer": "Yes"
      },
      {
        "augmented": true,
        "context": "There is a single person walking in the city that does not sleep and does not rest.",
        "question": "There is a single prostitute walking in the city that does not sleep and does not rest.",
        "reasoning": "The provided context states that there is a single person walking in the city who does not sleep and does not rest. However, the question specifically refers to this person as a prostitute, which is not mentioned or implied in the context. The context does not provide any information about the person's profession or identity beyond their behavior of not sleeping or resting. Therefore, there is no basis in the context to conclude that the person is a prostitute.",
        "answer": "No"
      },
      {
        "augmented": true,
        "context": "There is not a boat in this lake when it's not sunny.",
        "question": "There is not a dinghy in this lake when it's not sunny.",
        "reasoning": "The statement \"There is not a boat in this lake when it's not sunny\" implies that no boats are present in the lake unless it is sunny. A dinghy is a type of boat, so if no boats are present when it's not sunny, it logically follows that no dinghies would be present either.",
        "answer": "Yes"
      }
    ],
    "signature": {
      "instructions": "Analyze the given context and question using a step-by-step chain-of-thought reasoning approach to determine the logical consistency between them. Ensure to articulate your reasoning process clearly, identifying how each element of the context influences or relates to the question. Based on your logical analysis, provide a conclusive \"Yes\" or \"No\" answer indicating whether the question aligns with the context. Begin by evaluating the context, then the question, and finally synthesizing them together to reach your conclusion.",
      "fields": [
        {
          "prefix": "Context:",
          "description": "${context}"
        },
        {
          "prefix": "Question:",
          "description": "${question}"
        },
        {
          "prefix": "Reasoning: Let's think step by step in order to",
          "description": "${reasoning}"
        },
        {
          "prefix": "Answer:",
          "description": "Yes or No"
        }
      ]
    }
  }
}
\end{lstlisting}

\end{document}